\documentclass{article}

\usepackage{arxiv}
\usepackage[utf8]{inputenc}
\usepackage[T1]{fontenc}
\usepackage[pdftex, colorlinks=true, linkcolor=snsblue, citecolor=snsgreen, urlcolor=snsred]{hyperref}
\usepackage{url}
\usepackage{booktabs}
\usepackage{amsmath}
\usepackage{amssymb}
\usepackage{amsfonts}
\usepackage{nicefrac}
\usepackage{microtype}
\usepackage{natbib}
\usepackage{siunitx}
\usepackage{mathtools}
\usepackage{algorithm}
\usepackage{algorithmic}
\usepackage[nameinlink]{cleveref}

\usepackage{xcolor}
\definecolor{snsblue}{RGB}{76, 114, 176}
\definecolor{snsgreen}{RGB}{85, 168, 104}
\definecolor{snsred}{RGB}{196, 78, 82}
\definecolor{snspurple}{RGB}{129, 114, 178}
\definecolor{snsyellow}{RGB}{204, 185, 116}
\definecolor{snscyan}{RGB}{100, 181, 205}
\DeclareMathOperator{\RMSE}{RMSE}

\DeclareMathOperator{\FS}{FS}
\DeclareMathOperator{\intdiv}{{\mathbin{\!/\mkern-5mu/\!}}}

\title{A comparison of combined data assimilation and machine learning methods for offline and online model error correction}

\author{
Alban Farchi\\
CEREA, École des Ponts and EDF R\&D\\
\^Ile--de--France, France\\
\texttt{alban.farchi@enpc.fr}\\
\And
Marc Bocquet\\
CEREA, École des Ponts and EDF R\&D\\
\^Ile--de--France, France\\
\AND
Patrick Laloyaux\\
ECMWF\\
Shinfield Park\\
Reading, United Kingdom\\
\And
Massimo Bonavita\\
ECMWF\\
Shinfield Park\\
Reading, United Kingdom\\
\AND
Quentin Malartic\\
CEREA, École des Ponts and EDF R\&D\\
\^Ile--de--France, France\\
and\\
LMD/IPSL École Normale Supérieure and PSL University, École Polytechnique,\\
Université Paris-Saclay, Sorbonne Université, CNRS\\
Paris, France
}

\begin{document}

\maketitle

\begin{abstract}
Recent studies have shown that it is possible to combine machine learning methods with data assimilation to reconstruct a dynamical system using only sparse and noisy observations of that system. The same approach can be used to correct the error of a knowledge-based model. The resulting surrogate model is hybrid, with a statistical part supplementing a physical part. In practice, the correction can be added as an integrated term (\textit{i.e.} in the model resolvent) or directly inside the tendencies of the physical model. The resolvent correction is easy to implement. The tendency correction is more technical, in particular it requires the adjoint of the physical model, but also more flexible. We use the two-scale Lorenz model to compare the two methods. The accuracy in long-range forecast experiments is somewhat similar between the surrogate models using the resolvent correction and the tendency correction. By contrast, the surrogate models using the tendency correction significantly outperform the surrogate models using the resolvent correction in data assimilation experiments. Finally, we show that the tendency correction opens the possibility to make online model error correction, \textit{i.e.} improving the model progressively as new observations become available. The resulting algorithm can be seen as a new formulation of weak-constraint 4D-Var. We compare online and offline learning using the same framework with the two-scale Lorenz system, and show that with online learning, it is possible to extract all the information from sparse and noisy observations.
\end{abstract}


\section{Introduction: machine learning for model error correction}

Over the past decade, data-driven methods, and in particular machine learning (ML), have shown remarkable success in reproducing complex spatiotemporal processes, and have therefore been used in an increasing number of applications \citep{lecun-2015, goodfellow-2016, chollet-2018}. In the geosciences only, there is a fairly recent wealth of studies dealing with the problem of inferring the dynamics of a system from observations. Typical examples include the use of analogs, delay coordinates embedding, random forests, echo state networks and other neural networks such as residual, recurrent, or convolutional neural networks \citep{brunton-2016, hamilton-2016, lguensat-2017, pathak-2018, dueben-2018, fablet-2018, scher-2019, weyn-2019, arcomano-2020}. Most, if not all, of these examples implement a type of supervised learning where the goal is to minimise the loss function, a measure of the discrepancy between the statistical model (also called surrogate model) predictions and the observation dataset. The underlying assumption is that the system is fully observed without or with very little noise. In order to handle sparse and noisy observations, which is the case in most realistic systems in the geosciences, more and more studies consider the possibility of hybridising ML and data assimilation (DA) techniques \citep{abarbanel-2018, bocquet-2019a, brajard-2020, bocquet-2020, arcucci-2021}. In practice, DA tools are used, with the surrogate model, to estimate the state of the system from the observations while ML tools are used to estimate the surrogate model from the analysis (estimated) state. This method has been reformulated using a unifying Bayesian formalism by \citet{bocquet-2020}.

In the geosciences, even though models are affected by errors (\textit{e.g.}, misrepresented physical phenomena, unresolved small-scale processes, numerical integration errors, etc), they benefit from a long history of modelling and therefore they already provide a solid baseline. For this reason, recent studies focus on using ML techniques for model error correction instead of full model emulation \citep{rasp-2018, bolton-2019, jia-2019, watson-2019, bonavita-2020, brajard-2020b, gagne-2020, wikner-2020, farchi-2021}. The idea is to build a hybrid model with a physical, knowledge-based part, and a statistical part to supplement it. This means that the statistical model is trained to learn the error of the physical model. The underlying rationale is that model error correction should be an easier inference problem than full model emulation \citep{jia-2019, watson-2019, farchi-2021}.

From a technical perspective, the geoscientific models are based on a set of physical laws, usually represented as ordinary or partial differential equations (ODEs or PDEs). These equations define the \emph{tendencies} of the model. A numerical scheme is used to integrate them for a small time step, and several integration steps are composed to define the \emph{resolvent} between two forecast times. Following \citet{farchi-2021}, two strategies are possible for a correction term: (i) apply an integrated correction between two forecast times, \textit{i.e.} in the resolvent, or (ii) apply a correction directly in the tendencies. The first method is by far the simplest to implement, which is why it is the most widely applied, but it faces some limitations, in particular when using the hybrid model for DA experiments. The first objective of the present paper is hence to make an exhaustive comparison of the two methods for both forecast and assimilation experiments in a simplified modelling framework.

Beyond the design of the model error correction --- or more generally of any surrogate model --- the question of the use of observations arise. In most cases, the statistical model is only trained once the entire observation dataset is available: this is called \emph{offline learning}. The other option, \emph{online}, or \emph{sequential learning}, \textit{i.e.} improving the surrogate model as new observations become available, is also possible in ML, even if it is less common because the methods usually require very large datasets to achieve good performance. In a context where information is only available through sparse and noisy observations, this means that we have to learn both the state of the system and the surrogate model at the same time. This is the topic of several recent studies \citep{bocquet-2020a, gottwald-2021}, which emphasise the connections between this problem and classical parameter estimation in DA \citep{ruiz-2013, pulido-2018}. In the geosciences, online learning is more natural because observations are acquired sequentially, and improvements can be expected before having a long series of observations since the training begins from the first observation. Therefore, the second objective of the present paper is to explore the possibility to use online learning for model error correction.

The paper is organised as follows. \Cref{sec:methodology-offline} introduces the main methodological aspects for offline learning. We start with a brief overview of the Bayesian framework for combining DA and ML and how it can be used for model error correction. We then discuss the advantages and drawbacks of applying a correction term in the resolvent or in the tendencies, with an emphasis on the implications for forecast and assimilation applications. The two methods are compared in \cref{sec:illustration-offline} using the two-scale Lorenz model \citep[L05III,][]{lorenz-2005}. \Cref{sec:methodology-online} further develops the methodology to enable online learning for model error correction. \Cref{sec:illustration-online} illustrates the use of online learning with the same L05III model, and compares it to offline learning. Finally, conclusions are drawn in \cref{sec:conclusions}.

\section{Offline learning of model error with resolvent or tendency correction}
\label{sec:methodology-offline}

\subsection{A Bayesian framework for data assimilation and machine learning}
\label{ssec:methodology-offline-bayesian-framework-da-ml}

The starting point of the present work is a series of observations $\mathbf{y}_{k}\in\mathbb{R}^{N_{\mathsf{y}}}$ of a system at discrete times $t_{k}$ for $k\in\mathbb{N}$. The state of the system is represented by a vector $\mathbf{x}_{k}\in\mathbb{R}^{N_{\mathsf{x}}}$. The observations are related to the state through the observation equation
\begin{equation}
    \mathbf{y}_{k} = \mathcal{H}_{k}\left(\mathbf{x}_{k}\right) + \mathbf{v}_{k},
\end{equation}
where $\mathcal{H}_{k}:\mathbb{R}^{N_{\mathsf{x}}}\to\mathbb{R}^{N_{\mathsf{y}}}$ is the observation operator and $\mathbf{v}_{k}\in\mathbb{R}^{N_{\mathsf{y}}}$ the observation error at time $t_{k}$. We assume that the time evolution of the state is governed by the state equation
\begin{equation}
    \mathbf{x}_{k+1} = \mathcal{M}^{\mathsf{t}}_{k}\left(\mathbf{x}_{k}\right) + \mathbf{w}_{k},
\end{equation}
where $\mathcal{M}^{\mathsf{t}}_{k}:\mathbb{R}^{N_{\mathsf{x}}}\to\mathbb{R}^{N_{\mathsf{x}}}$ is the resolvent of the (unknown) true dynamical model from $t_{k}$ to $t_{k+1}$, and $\mathbf{w}_{k}\in\mathbb{R}^{N_{\mathsf{x}}}$ is the corresponding model error (\textit{e.g.}, related to sub-scale processes). To simplify the presentation, we make the following assumptions:
\begin{itemize}
    \item observations are available at regular intervals $t_{k}=k\Delta t$;
    \item the observation operator is constant over time $\mathcal{H}_{k}\equiv\mathcal{H}$;
    \item the observation error is uncorrelated in time and normally distributed $\mathbf{v}_{k}\sim\mathcal{N}\left(\mathbf{0},\mathbf{R}\right)$, where $\mathbf{R}$ is the observation error covariance matrix;
    \item the model error $\mathbf{w}_{k}$ is uncorrelated to the observation error $\mathbf{v}_{k}$.
\end{itemize}
In particular, the third point implies that the observations are not biased, which helps to attribute correctly the model errors. Furthermore, we also make the assumption that the true dynamical model is autonomous, in which case $\mathcal{M}^{\mathsf{t}}_{k}\equiv\mathcal{M}^{\mathsf{t}}_{\Delta t}$ the resolvent of the surrogate model for a $\Delta t$ integration. The extension of the present work to non-autonomous dynamics is not trivial and briefly discussed in \cref{ssec:illustration-online-generalisation-non-autonomous}.

Our goal is to derive a surrogate of the true model, which can be used to predict $\mathbf{x}_{k+1}$ from $\mathbf{x}_{k}$. Let $\mathbf{p}$ be the set of parameters defining the surrogate model. The discrepancy between the surrogate model predictions and the observations is measured with a cost function. A traditional ML approach to this problem is to use dense observations (\textit{i.e.}, $\mathcal{H}=\mathbf{I}$ the identity operator) and to neglect the observation errors (\textit{i.e.}, assuming that $\mathbf{R}=\mathbf{0}$), which yields the following cost function:
\begin{equation}
    \label{eq:methodology-offline-cost-dense}
    \mathcal{J}\left(\mathbf{p}\right) \triangleq \mathcal{L}\left(\mathbf{p}\right) + \frac{1}{2}\sum_{k=0}^{N_{\mathsf{t}}-1} \big\|\mathbf{y}_{k+1}-\mathcal{M}_{\Delta t}\left(\mathbf{p}, \mathbf{y}_{k}\right)\big\|^{2}_{\mathbf{Q}^{-1}_{k}},
\end{equation}
where $\mathcal{L}$ is a regularisation (prior) term on $\mathbf{p}$, $N_{\mathsf{t}}$ is the number of observation batches used to define $\mathcal{J}$, and $\mathbf{x}\mapsto\mathcal{M}_{\Delta t}\left(\mathbf{p}, \mathbf{x}\right)$ is the resolvent of the surrogate model for a $\Delta t$ integration.
The matrix norm notation $\left\|\mathbf{v}\right\|^{2}_{\mathbf{A}}$ stands for $\mathbf{v}^{\top}\mathbf{Av}$, and $\mathbf{Q}_{k}$ is the model error covariance matrix at time $t_{k}$.

With sparse observations, the problem is more complex because in order to derive the surrogate model, we need to estimate the true state. A rigorous Bayesian approach to this problem consists in extending \cref{eq:methodology-offline-cost-dense} to include the system trajectory $\mathbf{x}_{0}, \ldots, \mathbf{x}_{N_{\mathsf{t}}}$ in the control variables \citep{hsieh-1998, abarbanel-2018, bocquet-2019a, bocquet-2020}. The joint cost function reads
\begin{equation}
    \label{eq:methodology-offline-cost-sparse}
    \mathcal{J}\left(\mathbf{p}, \mathbf{x}_{0}, \ldots, \mathbf{x}_{N_{\mathsf{t}}}\right) \triangleq \mathcal{L}\left(\mathbf{p}, \mathbf{x}_{0}\right) + \frac{1}{2}\sum_{k=0}^{N_{\mathsf{t}}-1} \big\|\mathbf{x}_{k+1}-\mathcal{M}_{\Delta t}\left(\mathbf{p}, \mathbf{x}_{k}\right)\big\|^{2}_{\mathbf{Q}^{-1}_{k}} + \frac{1}{2}\sum_{k=0}^{N_{\mathsf{t}}} \big\|\mathbf{y}_{k}-\mathcal{H}\left(\mathbf{x}_{k}\right)\big\|^{2}_{\mathbf{R}^{-1}},
\end{equation}
where $\mathcal{L}$ is a regularisation term on both $\mathbf{p}$ and $\mathbf{x}_{0}$. The second term in \cref{eq:methodology-offline-cost-sparse} corresponds to the second term in \cref{eq:methodology-offline-cost-dense}, in which the observations have been replaced with the state, and the third term is the observation error term. \Cref{eq:methodology-offline-cost-sparse} is overall very similar to a typical weak-constraint (WC) 4D-Var cost function \citep{tremolet-2006}.

\begin{figure}[tbh]
    \centering
    \includegraphics{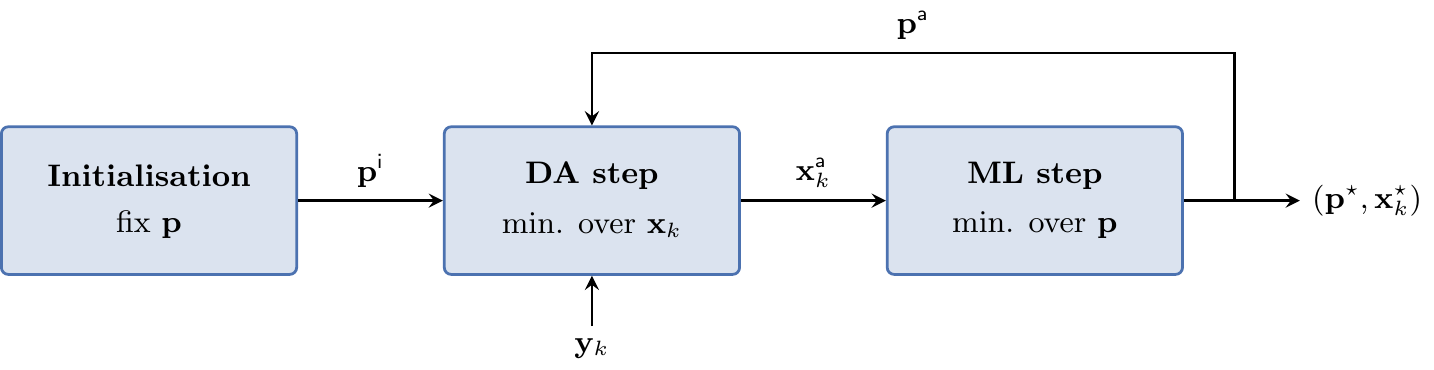}
    \caption{Illustration of the DA-ML method for the minimisation strategy of the cost function, \cref{eq:methodology-offline-cost-sparse}: alternate DA steps with ML steps to estimate the model parameters $\mathbf{p}$ and the state trajectory $\mathbf{x}_{0}, \ldots, \mathbf{x}_{N_{\mathsf{t}}}$ with an increasing accuracy.}
    \label{fig:methodology-offline-minimisation}
\end{figure}

Because the size of the trajectory control vector $N_{\mathsf{t}}\times N_{\mathsf{x}}$ is likely to be large, an efficient minimisation method relies on a coordinate descent technique, alternating DA steps to estimate the state with ML steps to estimate the surrogate model \citep{brajard-2020, bocquet-2020}. This combined DA-ML method, illustrated in \cref{fig:methodology-offline-minimisation}, explicitly exploits the different nature between the arguments of $\mathcal{J}$ (state of the system and surrogate model parameters) and is highly flexible since the DA and ML steps are independent. A comprehensive description of the DA-ML method is given by \citet{bocquet-2020}. 

The DA-ML method has first been used for full model emulation, \textit{e.g.} by \citet{brajard-2020}. In their example, the surrogate model is a neural network (NN) which represents the model tendencies. It is combined with an integration scheme to define the resolvent between two time steps. In this case, $\mathbf{p}$ corresponds to the set of weights and biases of the NN. The method has then been used to correct an imperfect physical model by \citet{brajard-2020b, farchi-2021}. For this problem, the formalism is simply obtained by replacing the resolvent of the surrogate model $\mathcal{M}_{\Delta t}$ with the resolvent of the corrected model, in particular in \cref{eq:methodology-offline-cost-sparse}. In general, model error correction should be an easier inference problem than full model emulation, which means that smaller surrogate models (smaller in number of parameters) and less training data are necessary. Moreover, using a physical model is likely to be beneficial to the method, in particular during the DA steps. It also solves the issue of the initialisation: the first step of the method is to perform DA with the (non-corrected) physical model. The advantages of model error correction over full model emulation are further investigated in \citet{farchi-2021}. 

\subsection{A typical geophysical model architecture}

In the present work, we investigate model error correction with the DA-ML method. Before we introduce any model error correction, we need to discuss the characteristics of the model to correct. The geophysical models rely on physical laws, which most of the time take the form of ODEs or PDEs.

The core of a model is a numerical code computing the model \emph{tendencies} $\phi$, which are defined as a discretised version of the differential equations in $\mathbb{R}^{N_{\mathsf{x}}}$:
\begin{equation}
    \label{eq:methodology-offline-def-tendencies}
    \phi\left(\mathbf{x}\right) \triangleq \frac{\mathrm{d}\mathbf{x}}{\mathrm{d}t}.
\end{equation}
The model tendencies are integrated over time step $\delta t$ using a dedicated integration scheme, for example the explicit Euler scheme:
\begin{equation}
    \mathcal{I}\left(\mathbf{x}\right) \triangleq \mathbf{x} + \delta t \cdot \phi\left(\mathbf{x}\right),
\end{equation}
or more elaborate schemes such as Runge--Kutta methods. Finally, several integration steps are composed to define the \emph{resolvent}\footnote{The term \emph{resolvent} is usual in the context of integral or differential equations. The same operator is often called \emph{flow}, or \emph{flow map} in dynamical systems and \emph{propagator} in theoretical physics.} from one time step to the next:
\begin{equation}
    \label{eq:methodology-offline-def-resolvent}
    \mathcal{M}_{\Delta t} \left(\mathbf{x}\right) \triangleq \mathcal{I}\circ\ldots\circ\mathcal{I}\left(\mathbf{x}\right).
\end{equation}
Two strategies can be used to correct such physical model. The first one is to include a correction in the resolvent, \cref{eq:methodology-offline-def-resolvent}. This is called resolvent correction (RC). The other strategy is to include the correction directly in the differential equations, in other words in the model tendencies, \cref{eq:methodology-offline-def-tendencies} \citep{bocquet-2019a}. This is called tendency correction (TC). According to \citet{farchi-2021}, both strategies have advantages and drawbacks. Let us illustrate the difference using a simple univariate example.

\subsection{Resolvent or tendency correction in a simple univariate example}
\label{ssec:methodology-offline-univariate-example}

Suppose that we follow the evolution of a process $x\in\mathbb{R}$ over two $\delta t$-integration steps with the explicit Euler scheme. The true two-step resolvent is given by
\begin{equation}
    \mathcal{M}^{\mathsf{t}}_{2}\left(x\right) = x+\delta t\cdot f\left(x\right)+\delta t\cdot f\left\{x+\delta t\cdot f\left(x\right)\right\},
\end{equation}
where $f$ represents the true model tendencies. Our imperfect physical model has tendencies $g$ and a two-step resolvent given by
\begin{equation}
    \mathcal{M}^{\mathsf{p}}_{2}\left(x\right) = x+\delta t\cdot g\left(x\right)+\delta t\cdot g\left\{x+\delta t\cdot g\left(x\right)\right\}.
\end{equation}
To simplify the expressions, in the following we take $\delta t=1$.

When using a TC, we assume that the corrected model has tendencies $g+\alpha$, whereas when using RC, we assume that the two-step resolvent of the corrected model is $\mathcal{M}^{\mathsf{p}}_{2}+\beta$. The optimal $\alpha$ and $\beta$ corrections are given by
\begin{align}
    \label{eq:methodology-offline-univariate-example-tc-optimal}
    \alpha^{\star}\left(x\right) &= f\left(x\right)-g\left(x\right),\\
    \beta^{\star}\left(x\right) &= \mathcal{M}^{\mathsf{t}}_{2}\left(x\right)-\mathcal{M}^{\mathsf{p}}_{2}\left(x\right)\\
    \label{eq:methodology-offline-univariate-example-rc-optimal}
    &= f\left(x\right)-g\left(x\right)+\textcolor{snsred}{f\left\{x+f\left(x\right)\right\}-g\left\{x+g\left(x\right)\right\}},
\end{align}
where the difference is highlighted in red. Obviously, the optimal $\beta$ is likely to be more complex than the optimal $\alpha$. The expression suggests that it will also be more nonlinear if $f$ or $g$ (or both) are nonlinear.

To further understand the difference, we derive the two-step resolvent with RC and TC, respectively written $\mathcal{M}^{\beta}_{2}$ and $\mathcal{M}^{\alpha}_{2}$:
\begin{align}
    \mathcal{M}^{\beta}_{2}\left(x\right) &= x+g\left(x\right)+g\left\{x+g\left(x\right)\right\}+\beta\left(x\right) \\
    \mathcal{M}^{\alpha}_{2}\left(x\right) &= x+g\left(x\right)+g\left\{x+g\left(x\right)+\textcolor{snsred}{\alpha\left(x\right)}\right\}+\alpha\left(x\right)+\textcolor{snsred}{\alpha\left\{x+g\left(x\right)+\alpha\left(x\right)\right\}},
\end{align}
where the difference is highlighted in red. From this perspective, it is clear that with TC, $\mathcal{M}^{\alpha}_{2}\left(x\right)$ is marked by the interaction between the physical model and the correction term $\alpha$. While this interaction is beneficial because it enhances $\mathcal{M}^{\alpha}_{2}\left(x\right)$, the downside is that inferring $\alpha$ from data is technically more difficult than inferring $\beta$. Let us see why.

Suppose that both $\alpha$ and $\beta$ depend on a coefficient $p\in\mathbb{R}$. Observation data usually come in the form of pairs $\left(x_{0}, x_{2}\right)$ with $x_{2}=\mathcal{M}^{\mathsf{t}}_{2}\left(x_{0}\right)$, possibly with some observation noise. Therefore, a learning step based on some kind of gradient descent would required the gradient of the \emph{corrected} two-step resolvent with respect to p, which is given by
\begin{align}
    \label{eq:methodology-offline-gradient-beta}
    \frac{\partial\mathcal{M}^{\beta}_{2}}{\partial p}\left(x\right) &= \frac{\partial\beta}{\partial p}\left(x\right),\\
    \label{eq:methodology-offline-gradient-alpha}
    \frac{\partial\mathcal{M}^{\alpha}_{2}}{\partial p}\left(x\right) &= \frac{\partial\alpha}{\partial p}\left(x\right)\cdot \left[1+\textcolor{snsred}{g'\left\{x+g\left(x\right)+\alpha\left(x\right)\right\}+\frac{\partial\alpha}{\partial p}\left\{x+g\left(x\right)+\alpha\left(x\right)\right\}}\right],
\end{align}
where the difference is once again highlighted in red. In particular, it depends on $g'$, the derivative of $g$. The equivalent for a geophysical numerical model would be the tangent linear (TL) operator, which may be difficult to compute.

To summarise, compared to RC, TC is more difficult to program because the correction term ($\alpha$ in the present example) is intrusive, meaning that it requires to modify deeply the code of the physical model. It is also more difficult to train, as illustrated by the difference between \cref{eq:methodology-offline-gradient-beta} and \cref{eq:methodology-offline-gradient-alpha}. On the other hand, once it is implemented, the TC has the potential to yield richer dynamics through the interaction with the physical model. Furthermore, by construction the RC can only correct the two-step resolvent, while the TC can also correct the one-step resolvent. This would make a difference when using the corrected model in a DA experiment with observations at every step, because then the one-step resolvent is explicitly needed. The simplest workaround is to assume a linear growth of errors in time \citep{brajard-2020b,farchi-2021}. In this case, the one-step resolvent with RC would be given by
\begin{equation}
    \mathcal{M}^{\beta}_{1}\left(x\right) = x+g\left(x\right)+\frac{1}{2}\beta\left(x\right),
\end{equation}
since $\beta$ is the correction term for the two-step resolvent $\mathcal{M}^{\beta}_{2}\left(x\right)$. However, even with the optimal $\beta$ correction from \cref{eq:methodology-offline-univariate-example-rc-optimal}, this one-step resolvent would still differ from the true one-step resolvent, given by
\begin{equation}
    \mathcal{M}^{\mathsf{t}}_{1}\left(x\right) = x+f\left(x\right).
\end{equation}
In their experiments, \citet{farchi-2021} found that this hypothesis was the main limitation for improving the accuracy of DA experiments with the corrected model. They concluded that the best strategy to correct a model to be used in DA experiments would probably be the TC.

Finally, let us mention that the model error correction considered in this section is autonomous and additive. The autonomous hypothesis can be relaxed, for example by including time in the set of predictors. However, one must keep in mind that in this case, the training dataset should capture the time evolution of the model error. The additive hypothesis can also be relaxed. Without prior knowledge on the model error form, using an additive correction is the simpler option but other choices are possible, \textit{e.g.} a multiplicative correction. Also note that if the physical model explicitly depends a set of parameters, the same framework can be used to calibrate these parameters.

\subsection{Comparing resolvent and tendency correction}
\label{ssec:methodology-offline-comparing-rc-tc}

\citet{farchi-2021} chose to focus on the RC because it is easier to implement. In the following section, we illustrate the difference between RC and TC. 
First hints in favour of the TC approach were gained from the comparison of the results of \citet{bocquet-2019a} and of \citet{brajard-2020} on the Lorenz 40-variable model.
To address the inference problem, we use the combined DA-ML method described in \cref{ssec:methodology-offline-bayesian-framework-da-ml}. We start by a DA step with an imperfect physical model to assimilate the observations. We then use a ML step to train a model error correction from the analysis of the DA step. Pushing the DA-ML method further, we could iterate in place: use the corrected model to get a more accurate analysis in further DA steps and learn from this more accurate analysis to get an improved model error correction in further ML steps, as illustrated by \cref{fig:methodology-offline-minimisation}. 

However, we choose to stop after the first DA-ML cycle for two reasons. First, DA experiments with realistic models are numerically expensive, and it may not be realistic to perform more than one DA step if the size of the trajectory $N_{\mathsf{t}}$ is large. It is also worth noting that operational centres usually compute \emph{reanalyses}, which means that the first DA step is a product which is likely to be already available \citep{hersbach-2020}. Second, if the physical model (without correction) is reasonably accurate, the analysis of the first DA step should be reasonably accurate and hence the improvement of the first ML step should be much larger than the improvement of further ML steps. However, even though we stop after the first DA-ML cycle, we perform a second DA step, but only for evaluation purposes.

Finally, we emphasise again the offline nature of the DA-ML method previously discussed. As is, the ML step starts only when the DA analysis is available, \textit{i.e.} once the entire observation dataset has been assimilated in the DA step. An alternative, online approach is proposed in \cref{sec:methodology-online}.

\section{Numerical illustration with the two-scale Lorenz model (part I)}
\label{sec:illustration-offline}

\subsection{Models description}
\label{ssec:illustration-offline-models-description}

In our experiments, the true model is the L05III model, which describes the evolution of two sets of variables: the slow variables $x_{n}$ for $n\in\left\{1,\ldots,N_{\mathsf{x}}\right\}$ and the fast variables $u_{m}$ for $m\in\left\{1,\ldots,N_{\mathsf{x}}\times N_{\mathsf{u}}\right\}$. These two-scale dynamics are given by
\begin{subequations}
    \begin{align}
        \frac{\mathrm{d}x_{n}}{\mathrm{d}t} &= x_{n-1}\left(x_{n+1}-x_{n-2}\right)-x_{n}+F-\frac{hc}{b}\sum_{m=1}^{N_{\mathsf{u}}}u_{m+\left(n-1\right)N_{\mathsf{u}}},\\
        \frac{\mathrm{d}u_{m}}{\mathrm{d}t} &= \frac{c}{b}\left\{b^{2}u_{m+1}\left(u_{m-1}-u_{m+2}\right)-bu_{m}\right\} + \frac{hc}{b}x_{1+\left(m-1\right)\intdiv N_{\mathsf{u}}},
    \end{align}
\end{subequations}

where $\intdiv$ is the integer division and where the indices are applied periodically: $x_{N_{\mathsf{x}}+n}=x_{n}$ and $u_{N_{\mathsf{x}}\times N_{\mathsf{u}}+m}=u_{m}$. The idea is that each slow variable $x_{n}$ is coupled to the $N_{\mathsf{u}}$ fast variables $u_{m}$ for  $m\in\left\{1+\left(n-1\right)N_{\mathsf{u}},\ldots,nN_{\mathsf{u}}\right\}$.

A first order approximation of the L05III model is the one-scale Lorenz model \citep[L96,][]{lorenz-1998}, which only describes the evolution of the slow variables $x_{n}$. The model is defined by
\begin{equation}
    \label{eq:illustration-offline-l96-model-ode}
    \frac{\mathrm{d}x_{n}}{\mathrm{d}t} = x_{n-1}\left(x_{n+1}-x_{n-2}\right)-x_{n}+F,
\end{equation}
where the indices once again apply periodically: $x_{N_{\mathsf{x}}+n}=x_{n}$. This model is used in our experiments as the (imperfect) physical model to correct.

Both L05III and L96 models are integrated using a fourth-order Runge--Kutta scheme, and the parameter values are reported in \cref{tab:illustration-offline-models-parametrisation}. With this setup, the true model dynamics is chaotic, with a leading Lyapunov exponent of $\num[round-mode=figures, round-precision=5]{1.3775}$ \citep{mitchell-2015} and the model variability, defined as the standard deviation of the climatological distribution of the state, averaged over the slow variables, is $\num[round-mode=figures, round-precision=5]{3.5371861209967395}$. When using the L96 model in place of the L05III model, two sources of model error are introduced:
\begin{enumerate}
    \item the fast variables $u_{m}$ generate unresolved processes;
    \item the integration time step $\delta t$ is $\num{0.05}$ instead of $\num{0.005}$.
\end{enumerate}
Moreover, even though the forcing coefficient $F$ differs in both models, this cannot strictly be considered as a third source of model error as $F=10$ is chosen for the L05III model to better match the dynamics of the L96 model with $F=8$.

\begin{table}[tbh]
    \centering
    \caption{Parametrisation for the true (L05III) and physical (L96) models.}
    \label{tab:illustration-offline-models-parametrisation}
    \begin{tabular}{lrrr} \toprule
    Parameter & Symbol & L05III & L96 \\ \midrule
    Number of slow variables & $N_{\mathsf{x}}$ & $\num{36}$ & $\num{36}$ \\
    Number of fast variables per slow variable & $N_{\mathsf{u}}$ & $\num{10}$ & \\
    Forcing & $F$ & $\num{10}$ & $\num{8}$ \\
    Coupling & $h$ & $\num{1}$ & \\
    Time-scale ratio & $c$ & $\num{10}$ & \\
    Space-scale ratio & $b$ & $\num{10}$ & \\
    Integration time step & $\delta t$ & $\num{0.005}$ & $\num{0.05}$ \\ \bottomrule 
    \end{tabular}
\end{table}

\begin{figure}[tbh]
    \centering
    \includegraphics{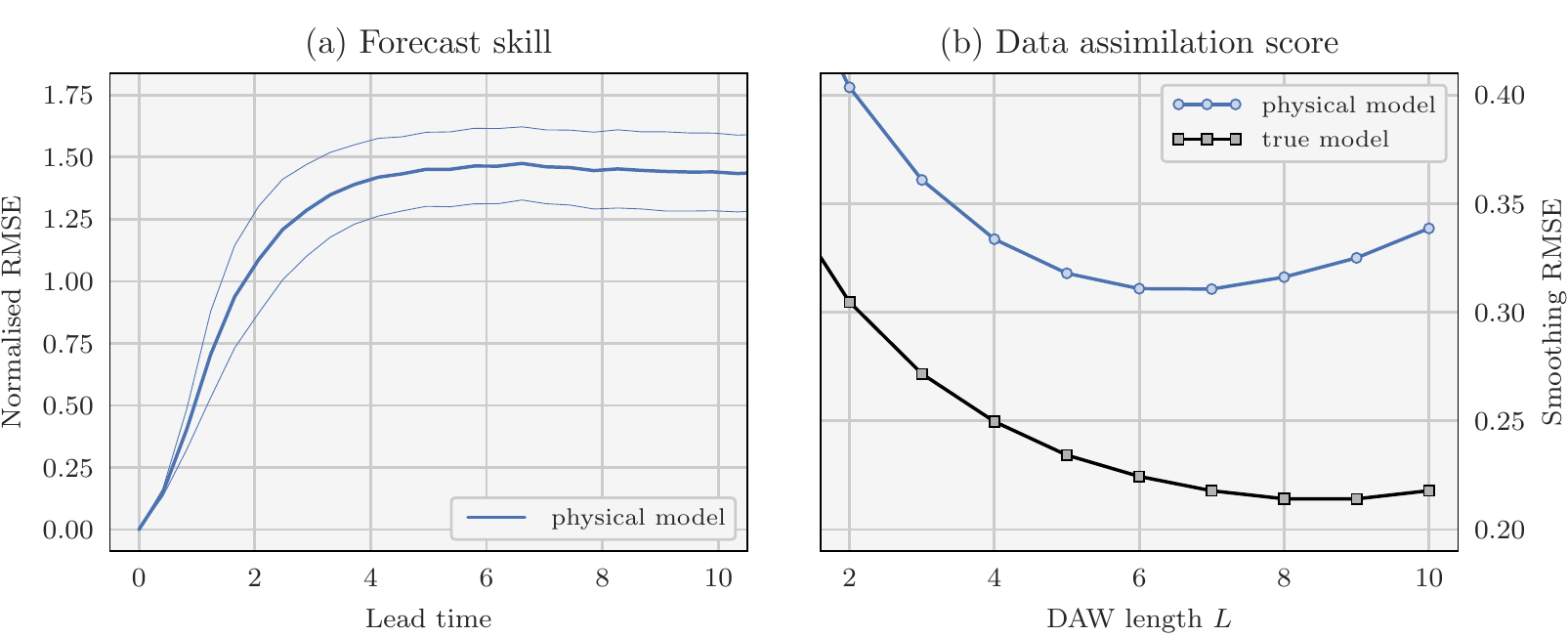}
    \caption{Left panel (a): forecast skill of the physical model (in units of the model variability) as a function of the lead time (in units of the Lyapunov time). The thick line shows the average over the $N_{\mathsf{e}}=\num{1024}$ initial conditions and the thin lines indicate plus or minus one standard deviation. Right panel (b): accuracy of the DA step as a function of the length of the DAW $L$ with the physical model (in blue) and with the true model (in black). The sRMSE is averaged over at least $\num{8192}\intdiv L$ cycles after a spin-up period of at least $\num{1024}\intdiv L$ cycles, and over $\num{16}$ repetitions of each experiment. For each value of $L$, $b$ is optimally tuned to yield the lowest sRMSE.}
    \label{fig:illustration-offline-models-fs-da}
\end{figure}

The accuracy of the physical (L96) model in reproducing the dynamics of the true (L05III) model is measured using the forecast skill (FS) defined as the average root-mean-squared error (RMSE) of the forecast after a given lead time:
\begin{equation}
    \FS\left(k\delta t\right) \triangleq \frac{1}{N_{\mathsf{e}}}\sum_{i=1}^{N_{\mathsf{e}}}\RMSE\left[\boldsymbol{\Pi}\circ\mathcal{M}^{\mathsf{t}}_{k \delta t}\left(\mathbf{x}_{i}, \mathbf{u}_{i}\right),\mathcal{M}^{\mathsf{p}}_{k \delta t}\left(\mathbf{x}_{i}\right)\right].
\end{equation}
In this equation, $\mathcal{M}^{\mathsf{t}}_{k \delta t}$ and $\mathcal{M}^{\mathsf{p}}_{k \delta t}$ are the resolvents of the true and physical models for a $k \delta t$ integration, respectively, $\boldsymbol{\Pi}$ is the projection operator onto the set of slow variables $\boldsymbol{\Pi}\left(\mathbf{x},\mathbf{u}\right)=\mathbf{x}$, and $\left(\mathbf{x}_{i}, \mathbf{u}_{i}\right)$ for  $i\in\left\{1,\ldots,N_{\mathsf{e}}\right\}$ is a set of $N_{\mathsf{e}}$ initial conditions representative of the true model climatology. The FS, normalised by the model variability, is shown in \cref{fig:illustration-offline-models-fs-da}a and illustrates the poor accuracy of the physical model. In the following sections, we will see how model error corrections can be used to improve the FS, but one must keep in mind that there is an intrinsic limit to potential improvements, because it is presumably impossible to exactly reproduce the dynamics of the true model with only $N_{\mathsf{x}}=\num{36}$ variables.

\subsection{Data assimilation with the physical model}
\label{ssec:illustration-offline-da-physical-model}

The first step of the DA-ML method is to perform DA with the physical model. The truth $\left(\mathbf{x}^{\mathsf{t}}_{k},\mathbf{u}^{\mathsf{t}}_{k}\right)$ is generated using the true model. Observations are taken every $\Delta t=\num{0.05}$ from the slow variables only, using 
\begin{equation}
    \label{eq:illustration-offline-da-observations}
    \mathbf{y}_{k} = \mathbf{x}^{\mathsf{t}}_{k} + \mathbf{v}_{k}, \quad \mathbf{v}_{k}\sim\mathcal{N}\left(\mathbf{0},\mathbf{I}\right).
\end{equation}
In other words, the observation operator is $\mathcal{H}=\mathbf{I}$, the observations are not biased, and the observation error covariance matrix is $\mathbf{R}=\mathbf{I}$. Numerical illustrations with sparse observation operators are provided, \textit{e.g.}, by \citet{brajard-2020, bocquet-2020, brajard-2020b, farchi-2021}. Except in the case of very sparse observations, the results are qualitatively similar over a wide range of observation density, which is why, for the present study, we have chosen to use dense observations for simplicity.

As explained in \cref{ssec:methodology-offline-bayesian-framework-da-ml}, the goal of the DA step is to minimise \cref{eq:methodology-offline-cost-sparse} with respect to the state trajectory $\mathbf{x}_{0}, \ldots, \mathbf{x}_{N_{\mathsf{t}}}$, which is a WC 4D-Var problem. However, solving a WC minimisation is probably unaffordable for large trajectories, as discussed by \citet{bocquet-2020}. To overcome this issue, we choose to assimilate the observations using the cycled strong-constraint (SC) 4D-Var algorithm, with consecutive DA windows (DAWs) of $L$ batches of observations. This will provide an approximate solution to the WC 4D-Var problem. More specifically, each 4D-Var problem consists in minimising the cost function
\begin{equation}
    \label{eq:llustration-offline-da-setup-cost-function}
    \mathcal{J}\left(\mathbf{x}_{k}\right) = \frac{1}{2}\left\|\mathbf{x}_{k}-\mathbf{x}^{\mathsf{b}}_{k}\right\|^{2}_{\mathbf{B^{-1}}} + \frac{1}{2}\sum_{l=0}^{L-1}\left\|\mathbf{y}_{k+l}-\mathcal{M}^{\mathsf{p}}_{l\Delta t}\left(\mathbf{x}_{k}\right)\right\|^{2}_{\mathbf{R^{-1}}},
\end{equation}
where $\mathbf{x}^{\mathsf{b}}_{k}$ is the background, $\mathbf{B}$ is the background error covariance matrix, $\left\{\mathbf{y}_{k},\ldots,\mathbf{y}_{k+L-1}\right\}$ is the set of assimilated observations, and $\mathcal{M}^{\mathsf{p}}_{l\Delta t}$ is the resolvent of the physical model for an integration of $l\Delta t$. The analysis is performed at time $t_{k}$ (the time of the first batch of assimilated observations) and, in this cycled context, it is used to obtain the background for the next analysis which is performed at time $t_{k+L}$, using
\begin{equation}
    \mathbf{x}^{\mathsf{b}}_{k+L} = \mathcal{M}^{\mathsf{p}}_{L\Delta t}\left(\mathbf{x}^{\mathsf{a}}_{k}\right).
\end{equation}
For the first cycle, the background state is obtained by perturbing the truth:
\begin{equation}
    \mathbf{x}^{\mathsf{b}}_{0} = \mathbf{x}^{\mathsf{t}}_{0}+\mathbf{w}, \quad \mathbf{w}\sim\mathcal{N}\left(\mathbf{0},\mathbf{I}\right).
\end{equation}
Finally, the background error covariance matrix $\mathbf{B}$ is set to $b^{2}\mathbf{I}$, where $b$ is an algorithmic parameter to specify.

At each cycle, the cost function $\mathcal{J}$, \cref{eq:llustration-offline-da-setup-cost-function}, is minimised using the L-BFGS algorithm \citep{byrd-1995}, a quasi-Newton minimisation algorithm. The gradient of $\mathcal{J}$ is computed exactly using automatic differentiation, and the starting point of the minimisation is $\mathbf{x}^{\mathsf{b}}_{k}$. The accuracy of the DA step is measured using the RMSE of the analysis (analysis minus truth) at the start of the DAW, hereafter called the \emph{smoothing} RMSE (sRMSE), averaged over a sufficiently large number of cycles to ensure the convergence of the statistical indicators.

In order to choose an appropriate value for the length of the DAW $L$, we first study the evolution of the sRMSE as a function of $L$. The results are shown in \cref{fig:illustration-offline-models-fs-da}b. As expected, the sRMSE starts by decreasing with $L$. It reaches an optimum for $L=\num{6}$, and then increases with $L$ as the impact of model error grows. For comparison, \cref{fig:illustration-offline-models-fs-da} also shows the results when using the true model in place of the physical model. Note that in this case the 4D-Var cost function $\mathcal{J}$, \cref{eq:llustration-offline-da-setup-cost-function}, depends on both the slow and the fast variables $\mathbf{x}_{k}$ and $\mathbf{u}_{k}$. The evolution of the sRMSE as a function of $L$ is very similar, with the exception that the scores are overall much lower, and that the sRMSE increase for large values of $L$ does not come from model error but from optimisation issues. Indeed, for long DAWs, the cost function $\mathcal{J}$ is likely to have several local minima, which would make the L-BFGS algorithm not suited for the minimisation. Using a quasi-static formulation of 4D-Var could mitigate this issue \citep{pires-1996,fillion-2018}.

\subsection{Model error correction with a univariate polynomial regression}
\label{ssec:illustration-offline-regression}

The present model error setup, as described in \cref{ssec:illustration-offline-models-description}, has already been addressed outside the scope of ML, for example by \citet{wilks-2005}. The idea is to replace the physical model tendencies, \cref{eq:illustration-offline-l96-model-ode}, by
\begin{equation}
    \frac{\mathrm{d}x_{n}}{\mathrm{d}t} = x_{n-1}\left(x_{n+1}-x_{n-2}\right)-x_{n}+F+g\left(x_{n}\right),
\end{equation}
where $g$ is a univariate fourth-order polynomial correction, shared between all $N_{\mathsf{x}}=36$ slow variables. The five coefficients of $g$ are computed using a least-square regression of the difference between \cref{eq:illustration-offline-l96-model-ode} and the empirical tendencies 
\begin{equation}
    \frac{x^{\mathsf{t}}_{n}\left(t+\delta t\right)-x^{\mathsf{t}}_{n}\left(t\right)}{\delta t}
\end{equation}
computed from a trajectory $\mathbf{x}^{\mathsf{t}}\left(t\right)$ of the true model.

\begin{figure}[tbh]
    \centering
    \includegraphics{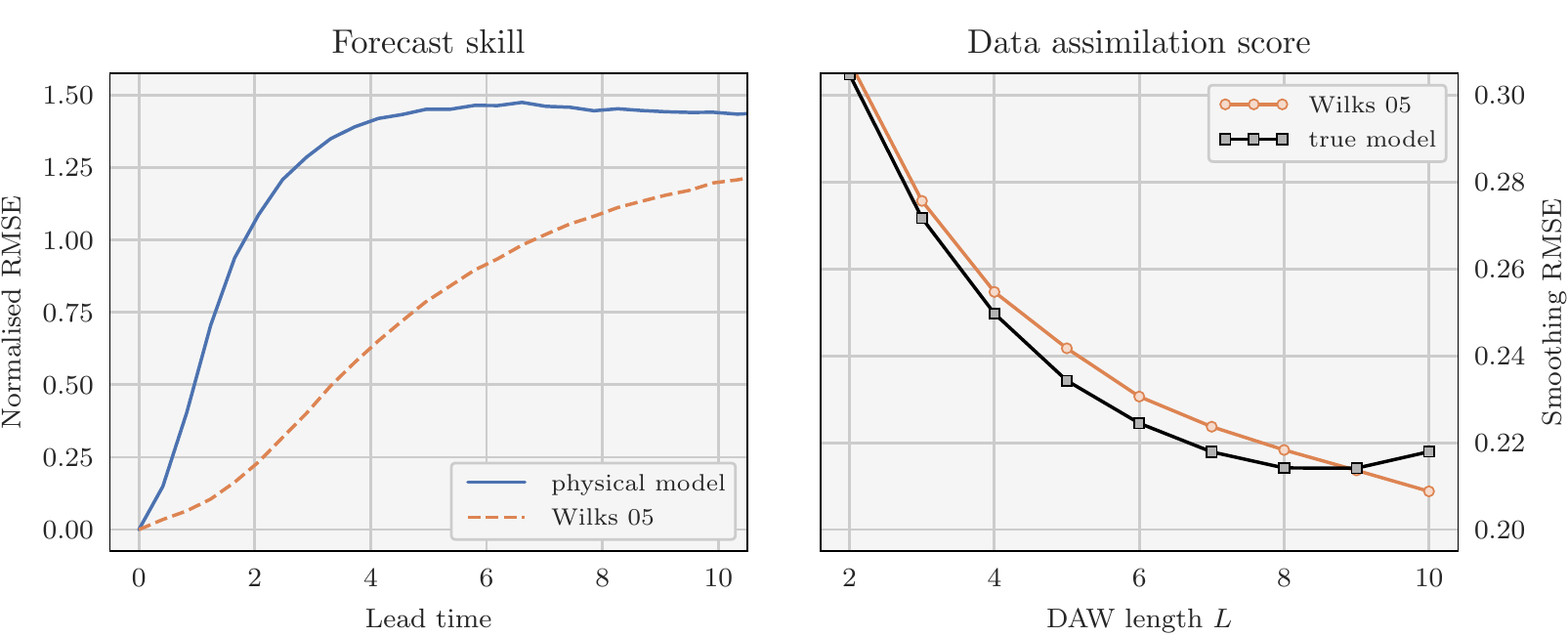}
    \caption{Same as \cref{fig:illustration-offline-models-fs-da} for the physical model (in blue), the model with polynomial regression (in orange) and the true model (in black).}
    \label{fig:illustration-offline-models-fs-da-regression}
\end{figure}

Offering a baseline score for later comparison, \cref{fig:illustration-offline-models-fs-da-regression} shows the FS and the DA score for the model with the polynomial regression $g$. For this illustration, following the approach of \citet{wilks-2005}, the coefficients of $g$ are computed using $\num{2000}$ pairs of snapshots $\left(\mathbf{x}^{\mathsf{t}}\left(t\right),\mathbf{x}^{\mathsf{t}}\left(t+\delta t\right)\right)$ with $\delta t=\num{0.005}$, the integration time step of the true model. The time interval between two consecutive pairs of snapshots is set to $\num{1000}$ integration steps. The results show that this simple correction is effective, both in forecast and DA experiments. In particular, the DA score is very close to the one obtained with the true model. It is even better for $L\geq\num{10}$. This probably comes from the fact that a small amount of model error regularises the cost function \cref{eq:llustration-offline-da-setup-cost-function} and mitigates the numerical issues discussed at the end of \cref{ssec:illustration-offline-da-physical-model}.

Of course, this method cannot be applied to realistic models because the regression requires the truth, at a very high frequency ($\delta t=0.005$). We have checked that using $\delta t=0.05$ (the integration time step of the physical model) yields an ineffective correction. Nevertheless, it shows that the model error structure in this setup can be effectively represented with a small number of parameters. The TC framework presented in \cref{sec:methodology-offline} can be seen as a generalisation of the method of \citet{wilks-2005} to (i) any kind of correction $g$, in particular multivariate ones, and (ii) sparse and noisy observations for the training. A more complex but less scalable model error correction scheme has also been proposed for the same model by \citet{pulido-2018}.

\subsection{Resolvent and tendency correction with the DA-ML method}
\label{ssec:illustration-offline-rc-tc}

\subsubsection{The data assimilation step}
\label{sssec:illustration-offline-rc-tc-da-step}

Given the results of \cref{ssec:illustration-offline-da-physical-model}, we start the DA-ML method by a long DA experiment with the physical model and with $L=6$. At each cycle, we only keep the analysis at the start of the DAW. The result is a time series of analysis snapshots $\mathbf{x}^{\mathsf{a}}_{kL}$, where the time interval between two snapshots is $L\Delta t=\num{0.3}$. This trajectory is used to build the training dataset for the ML step. In other words, the surrogate models are trained to reproduce the map
\begin{equation}
    \label{eq:illustration-offline-training-map}
    \mathbf{x}^{\mathsf{a}}_{kL} \mapsto \mathbf{x}^{\mathsf{a}}_{\left(k+1\right)L}.
\end{equation}
Another trajectory, resulting from a distinct long DA experiment, is used to build the validation dataset. Finally, since the ultimate goal is to predict the true dynamics and not the dynamics of the analysis snapshots, we compute an additional trajectory, this time with the true model. This third trajectory $\mathbf{x}^{\mathsf{t}}_{kL}$ is used to build the test dataset, and hence to evaluate the ability of the surrogate models to reproduce the map
\begin{equation}
    \label{eq:illustration-offline-def-map-truth}
    \mathbf{x}^{\mathsf{t}}_{kL} \mapsto \mathbf{x}^{\mathsf{t}}_{\left(k+1\right)L}.
\end{equation}

\subsubsection{Designing the surrogate models}
\label{sssec:illustration-offline-design-surr-models}

The second step of the DA-ML consists in defining and training a surrogate model with the analysis of the first DA step. In this section, three different surrogate models are tested to correct the physical model. All three of them are autonomous and use NNs. For the first surrogate, the correction is computed using a NN called CNN-a and then \emph{added to the resolvent} of the physical model, following the RC approach, which yields
\begin{equation}
    \mathcal{M}^{\mathsf{a}}_{L\Delta t}\left(\mathbf{p},\mathbf{x}\right) \triangleq \mathcal{M}^{\mathsf{p}}_{L\Delta t}\left(\mathbf{x}\right) + \mathcal{F}^{\mathsf{a}}\left(\mathbf{p},\mathbf{x}\right).
\end{equation}
In this equation, $\mathcal{M}^{\mathsf{p}}_{L\Delta t}$ is the resolvent of the physical model for an integration of $L\Delta t$ (one DAW), $\mathcal{F}^{\mathsf{a}}$ is the map encoding CNN-a, $\mathbf{p}$ is the set of parameters of CNN-a (the weights and biases of the NN), and $\mathcal{M}^{\mathsf{a}}_{L\Delta t}$ is the resolvent of the resulting surrogate model, called RC-CNN-a. For the second surrogate, the correction is computed using a NN called CNN-b and then \emph{added to the tendencies} of the physical model, following the TC approach:
\begin{equation}
    \phi^{\mathsf{b}}\left(\mathbf{p},\mathbf{x}\right) \triangleq \phi^{\mathsf{p}}\left(\mathbf{x}\right) + \mathcal{F}^{\mathsf{b}}\left(\mathbf{p},\mathbf{x}\right).
\end{equation}
In this equation, $\phi^{\mathsf{p}}$ represents the physical model tendencies, given by \cref{eq:illustration-offline-l96-model-ode}, $\mathcal{F}^{\mathsf{b}}$ is the function encoding CNN-b, $\mathbf{p}$ is the set of parameters of CNN-b, and $\phi^{\mathsf{b}}$ represents the tendencies of the resulting surrogate model, called TC-CNN-b. To compute the resolvent of this model, $\mathcal{M}^{\mathsf{b}}_{L\Delta t}$, we keep the integration scheme and time step of the physical model. Finally, the third surrogate model, called TC-CNN-c, is similar to TC-CNN-b with CNN-b replaced with another NN called CNN-c, which uses a different activation function.

As explained in \cref{ssec:illustration-offline-regression}, the model error structure is not overly complex. For this reason, we want to keep the NNs as simple as possible. We have experimented with several NNs configurations and have selected the following sequential (or feed-forward) architecture with:
\begin{enumerate}
    \item the input layer;
    \item a sequence of convolutional layers;
    \item a final convolutional layer as output layer (without activation).
\end{enumerate}
All intermediate convolutional layers share the same number of filters, the same convolutional window, and the same activation function. They also use periodic padding to preserve the input and output shape of the layers. The last convolutional layer uses only one filter, a convolution window of only one variable, and no activation function. The purpose of this layer is not to actually perform a convolution, but to project the output of the previous layer to the output variables. The settings of the intermediate convolutional layers are reported in \cref{tab:illustration-offline-cnn-parametrisation} for CNN-a, CNN-b, and CNN-c, alongside the total number of parameters.

\begin{table}[tbh]
    \centering
    \caption{Settings of the convolutional layers of the NNs used in the DA-ML method. The absence of an activation function is tantamount to a linear activation function.}
    \label{tab:illustration-offline-cnn-parametrisation}
    \begin{tabular}{lccc} \toprule
    Setting & CNN-a & CNN-b & CNN-c \\ \midrule
    Number of layers & $\num{4}$ & $\num{1}$ & $\num{1}$ \\
    Number of filters per layer & $\num{16}$ &  $\num{16}$ & $\num{16}$ \\
    Size of the convolutional window & $\num{5}$ &  $\num{5}$ & $\num{5}$ \\
    Activation function & $\tanh$ & & $\tanh$ \\ \midrule
    Total number of parameters & $\num{4001}$ & $\num{113}$ & $\num{113}$ \\\bottomrule
    \end{tabular}
\end{table}

\subsubsection{Neural networks initialisation}
\label{sssec:illustration-offline-nn-initialisation}

When working with NNs, the parameter initialisation step is important. A common method is to use random values for the initial weights and to set the initial biases to zero. The underlying idea is that there is no reason for the optimal weights to display any specific symmetry. Because such symmetries are preserved during the training, even with stochastic gradient descent, they need to be broken during the initialisation, hence the use of random initial values \citep{goodfellow-2016}.

In our case however, the situation is different because the surrogate models are hybrid. Since the corrections are additive, all three surrogate models are equivalent to the physical model when $\mathbf{p}=\mathbf{0}$, and it is highly probable that a random $\mathbf{p}$ would make the model predictions worse. Initialising the NNs with $\mathbf{p}=\mathbf{0}$ would hence make sense, but for the reasons aforementioned, this could yield suboptimal surrogate models after the training. Therefore, we initialise the NNs using the following approach, which we found to be a good compromise. The intermediate convolutional layers are initialised using the classical method in ML (random weights and zero biases) and the last convolutional layer is initialised to zero (both zero weights and zero biases). This approach is very similar to the ReZero method developed by \citet{bachlechner-2020}.

\subsubsection{Training the surrogate models}
\label{sssec:illustration-offline-training-surrogate-models}

We now start the ML step. The surrogate model parameters $\mathbf{p}$ are optimised using the Adam algorithm, a variant of the stochastic gradient descent \citep{kingma-2015}. The loss function is the mean-squared error (MSE) over the training dataset, made of analysis snapshots. The training consists of $\num{1024}$ epochs with a learning rate of $\num{1e-3}$ and a batch size of $\num{32}$. After the entire training step, we keep the model which yields the lowest MSE over the validation dataset, also made of analysis snapshots. This is necessary since the cost functions of the NNs do not include any internal mechanism to mitigate overfitting (\textit{e.g.} regularisation). Finally, we evaluate the trained model by computing the MSE over the test dataset, made of truth snapshots, hereafter called test MSE (tMSE). For comparison, we also train and evaluate the surrogate models using the exact same method but with snapshots from the truth (instead of the analysis) in the training and validation datasets. This is equivalent to using dense and noiseless observations.

\begin{figure}[tbh]
    \centering
    \includegraphics{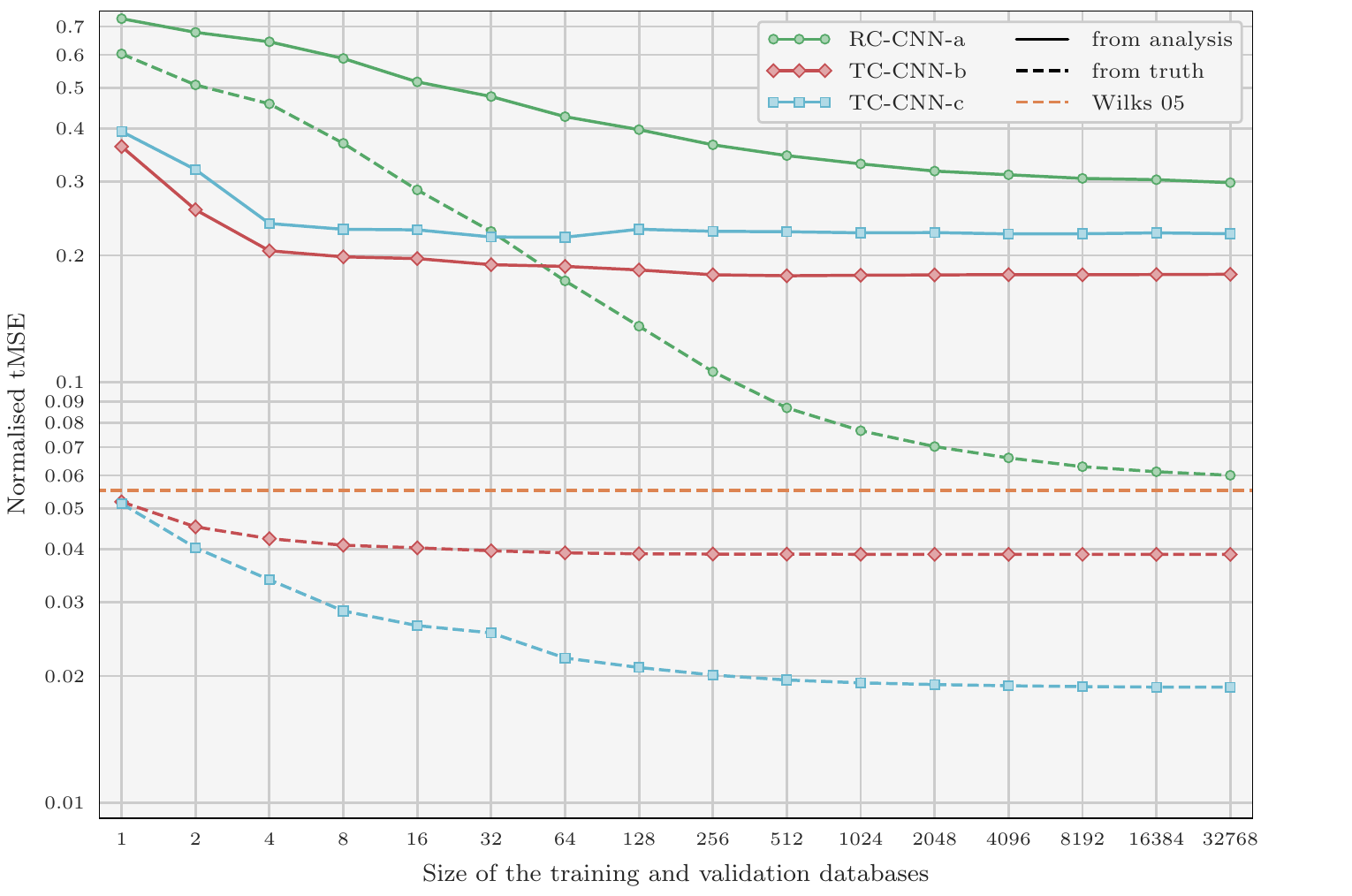}
    \caption{Evolution of the tMSE of the trained surrogate model as a function of the size of the training and validation datasets for RC-CNN-a (in green), TC-CNN-b (in red), and TC-CNN-c (in cyan). The tMSE is computed using a test dataset of size $N_{\mathsf{t}}=\num{8192}$ and then normalised by the tMSE of the physical model ($\num[round-mode=figures, round-precision=5]{0.2778511064813523}$). Each experiment (initialisation, training and evaluation) is repeated $\num{16}$ times with different training, validation, and test datasets and the final score is averaged over the $\num{16}$ repetitions. The surrogate models are trained either with the analysis (continuous lines) or with the truth (dashed lines). For comparison, the horizontal dashed orange line indicates the score for the model with the polynomial regression of \citet{wilks-2005}.}
    \label{fig:illustration-offline-models-tmse}
\end{figure}

\Cref{fig:illustration-offline-models-tmse} shows the training results for datasets of increasing size. Note that, in order to build a dataset of size $N_{\mathsf{t}}$, the total number of cycles (or DAWs) required in the preliminary DA step is $2\times\left(N_{\mathsf{t}}+1\right)$:
\begin{itemize}
    \item $N_{\mathsf{t}}+1$ cycles to produce the $N_{\mathsf{t}}$ pairs of analysis snapshots $\left(\mathbf{x}^{\mathsf{a}}_{kL},\mathbf{x}^{\mathsf{a}}_{\left(k+1\right)L}\right)$ for the training dataset;
    \item and the same for the validation dataset.
\end{itemize}
Let us first discuss the training with the truth, because it shows the full potential of each model. First, all three models do improve over the physical model and yield very low tMSEs, but the scores with TC are significantly better than with RC. As explained in \cref{ssec:methodology-offline-univariate-example}, the models with TC benefit from the interaction between the physical model and the correction term (the NN). This is even more important here than in the univariate example of \cref{ssec:methodology-offline-univariate-example}, because here the number of interactions for a prediction is $\num{24}$: $\num{4}$ interactions per integration step (through the fourth-order Runge--Kutta scheme) times $\num{6}$ integration steps between two DAWs. Furthermore, the training dataset needs to be much larger to get an accurate model with RC than with TC. This is consistent with the total number of trainable parameters for each model, reported in \cref{tab:illustration-offline-models-parametrisation}: $\num{4001}$ for RC-CNN-a and only $\num{113}$ for TC-CNN-b and TC-CNN-c. It is remarkable that with a training dataset of only one pair of truth snapshots (the smallest possible training dataset), the models with TC are already very accurate, more than RC-CNN-a trained with the largest dataset considered in the present study! Obviously, this is only possible because the correction is autonomous. The overall accuracy of TC-CNN-b is also remarkable, given the fact that the correction provided by CNN-b is linear. The only difference between CNN-b and CNN-c is their activation function for the intermediate layers. This means that the difference between TC-CNN-b and TC-CNN-c illustrates the nonlinearity of the error in the model tendencies: weak but non-negligible. For comparison, we have checked that with RC, the accuracy of the surrogate models built using linear NNs is not satisfactory. This shows that the nonlinearity of the dynamics over one DAW, \cref{eq:illustration-offline-def-map-truth}, is significant and that estimating and correcting model error as a tendency forcing is a good first-order strategy to mitigate the effects of nonlinearities. For completeness, we mention that better scores could have been obtained with RC, for example with larger or deeper NNs. However, this would increase the number of trainable parameter, which means that the training dataset should be even larger.

When training with the analysis, the accuracy of the surrogate models is much lower, which was expected, but all three models are still able to improve over the physical model. Most of the conclusions hold: the scores with TC are better than with RC and require much smaller datasets. However this time, the linear TC-CNN-b outperforms the nonlinear TC-CNN-c. This is probably a sign that nonlinear NNs are harder to train.

\subsubsection{Forecast skill of the surrogate models}

The tMSE is a measure of the accuracy of a model for an integration of one DAW of $L\Delta t=\num{0.3}$. In this section, we measure the accuracy of the surrogate models for longer forecast, by computing the FS as defined in \cref{ssec:illustration-offline-models-description}. In order not to penalise RC-CNN-a over TC-CNN-b and TC-CNN-c, we evaluate the surrogate models which have been trained with the largest dataset.

\begin{figure}[tbh]
    \centering
    \includegraphics{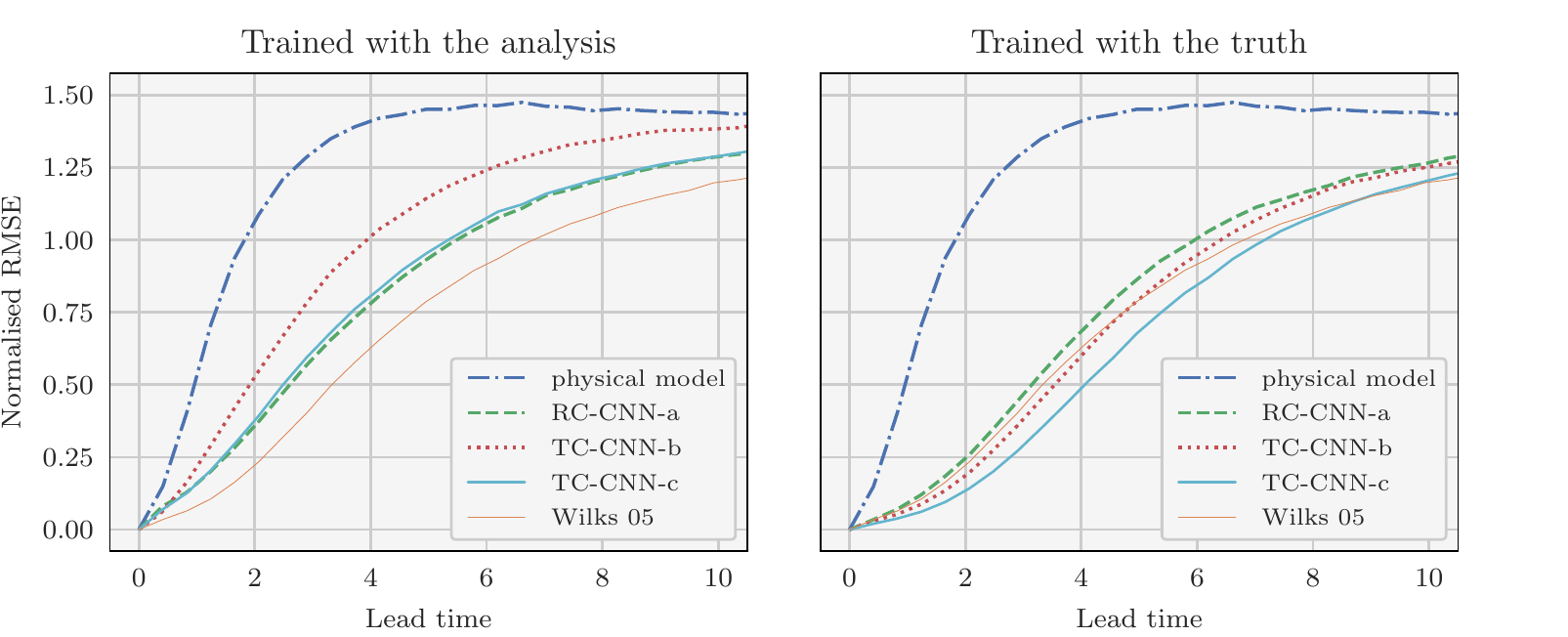}
    \caption{Same as \cref{fig:illustration-offline-models-fs-da}, left panel, for the physical model (in blue) and the trained surrogate models: RC-CNN-a (in green), TC-CNN-b (in red), and TC-CNN-c (in cyan). The surrogate models are trained either with the analysis (left panel) or with the truth (right panel). For comparison, the thin orange line indicates the scores for the model with the polynomial regression of \citet{wilks-2005}.}
    \label{fig:illustration-offline-fs}
\end{figure}

The results are shown in \cref{fig:illustration-offline-fs} and demonstrate that the models, trained for an integration of one DAW, remain effective for much longer integrations. When training with the truth, the ranking of the three surrogate models is clear and consistent with the tMSE results: TC-CNN-c is the most accurate, followed by TC-CNN-b, and the least accurate is RC-CNN-a. When training with the analysis, the ranking of the models for long integrations (\textit{e.g.} longer that $\num{6}$ DAWs) is the opposite of the ranking for the tMSE results: RC-CNN-a is the most accurate, very closely followed by TC-CNN-c, and the least accurate is TC-CNN-b, the only model built using a linear NN. There is a crossover in the forecast error curves after $\num{2}$ or $\num{3}$ DAWs, with the errors of the linear NN (TC-CNN-b) becoming larger than those of the nonlinear NNs (RC-CNN-a and TC-CNN-c). This issue is not specific to the L05III model nor to the use of nonlinear NNs since \citet{farchi-2021} also faced the same kind of issues with a quasi-geostrophic model and with linear NNs. We do not have yet a convincing explanation for this behaviour, which is specific to the use of the analysis for the training, and understanding it better will require further work. In any case, we conclude that, when the surrogate models are trained with the analysis, the accuracy for long forecasts is somewhat similar with RC and with TC and requires nonlinear corrections.

\subsubsection{Data assimilation experiments with the surrogate models}
\label{sssec:illustration-offline-corrected-da-exp}

In this section, the goal is to measure the accuracy of the surrogate models in DA experiments. To do that, we reimplement the 4D-Var setup described in \cref{ssec:illustration-offline-da-physical-model} replacing the physical model by one of the surrogate models. In particular, in the cost function \cref{eq:llustration-offline-da-setup-cost-function}, $\mathcal{M}^{\mathsf{p}}_{l\Delta t}$, the resolvent of the physical model for an integration of $l\Delta t$, is replaced with $\mathcal{M}^{\mathsf{a}}_{l\Delta t}$, $\mathcal{M}^{\mathsf{b}}_{l\Delta t}$, or $\mathcal{M}^{\mathsf{c}}_{l\Delta t}$, the resolvents of RC-CNN-a, TC-CNN-b, or TC-CNN-c for an integration of $l\Delta t$. While the construction of $\mathcal{M}^{\mathsf{b}}_{l\Delta t}$ and $\mathcal{M}^{\mathsf{c}}_{l\Delta t}$ is similar to that of $\mathcal{M}^{\mathsf{p}}_{l\Delta t}$ because TC-CNN-b and TC-CNN-c use the TC method, the construction of $\mathcal{M}^{\mathsf{a}}_{l\Delta t}$ requires an additional assumption because RC-CNN-a uses the RC method. As discussed in \cref{ssec:methodology-offline-univariate-example}, the simplest assumption is the linear growth of errors in time, for which 
\begin{equation}
    \mathcal{M}^{\mathsf{a}}_{\Delta t} \triangleq \mathcal{M}^{\mathsf{p}}_{\Delta t} + \frac{1}{L} \mathcal{F}^{\mathsf{a}},
\end{equation}
where $\mathcal{F}^{\mathsf{a}}$ is defined in \cref{sssec:illustration-offline-design-surr-models} as the function encoding CNN-a. This assumption is the standard assumption in the current implementation of WC 4D-Var \citep{laloyaux-2020, laloyaux-2020b} and it is the assumption we make for our DA experiments. Furthermore, for the same reason as above, we evaluate the models which have been trained with the largest dataset.

\begin{figure}[tbh]
    \centering
    \includegraphics{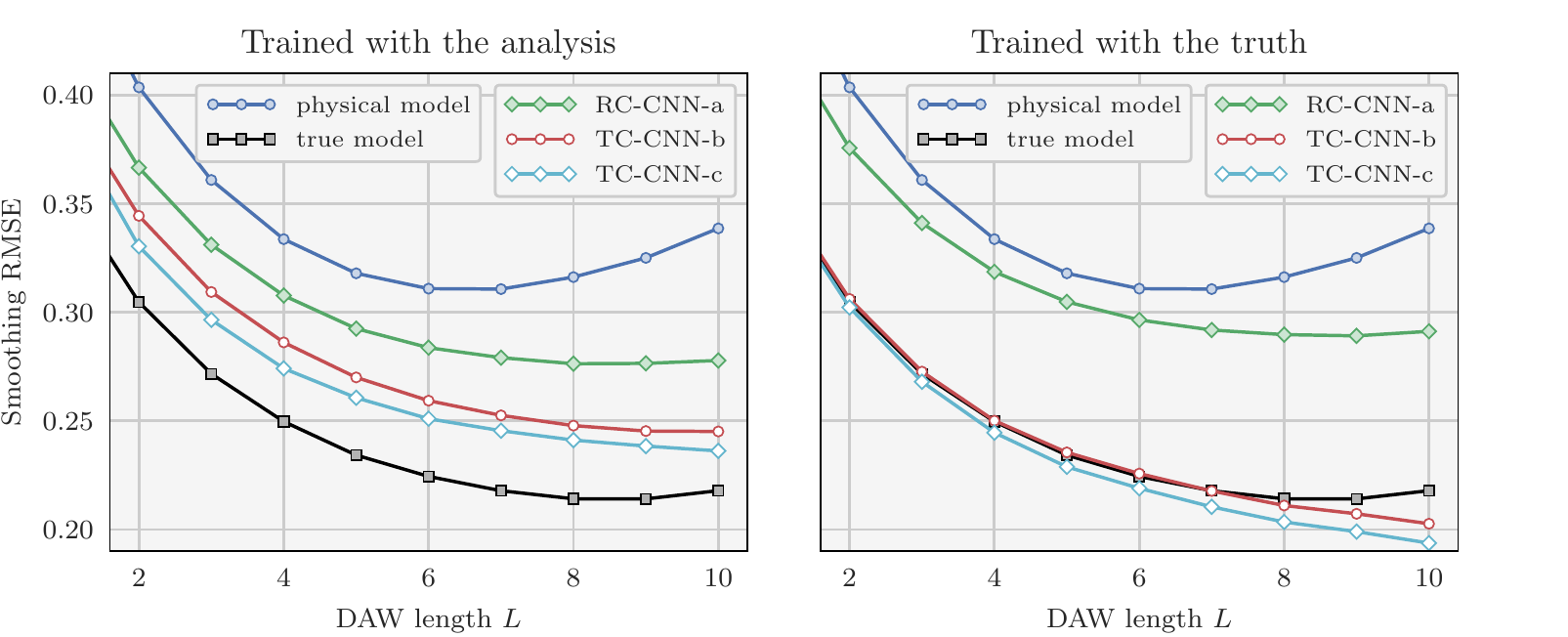}
    \caption{Same as \cref{fig:illustration-offline-models-fs-da}, right panel, for the physical model (in blue), the true model (in black), and the trained surrogate models: RC-CNN-a (in green), TC-CNN-b (in red), and TC-CNN-c (in cyan). The surrogate models are trained either with the analysis (left panel) or with the truth (right panel).}
    \label{fig:illustration-offline-da}
\end{figure}

Following the approach of \cref{ssec:illustration-offline-da-physical-model}, we study the evolution of the sRMSE as a function of the length of the DAW $L$. The results are shown in \cref{fig:illustration-offline-da} and demonstrate that the improved accuracy of the models also yields more accurate analyses. From these results, two elements could be highlighted. First, the sRMSE is much lower with TC-CNN-b and TC-CNN-c, which both use the TC method, than with RC-CNN-a, which uses the RC method. Additionally, the sRMSE is lower when RC-CNN-a is trained with the analysis than with the truth, even though the model error predictions are more accurate, as indicated by the lower tMSE. These results confirm the hypothesis of \citet{farchi-2021} that the main obstacle to more accurate analyses with the RC method is the assumption of linear growth of errors in time and that the TC method is more appropriate for DA experiments where the impact of nonlinearities is significant. Second, when trained with the truth, the sRMSE obtained with the TC method is of the same order as that obtained with the true model, it is even better for TC-CNN-c! For large windows, typically $L\geq\num{8}$, a fraction of the improvement comes from the numerical issues with the true model discussed at the end of \cref{ssec:illustration-offline-da-physical-model}. For smaller windows however, we believe that the remaining improvement shows that TC-CNN-c may capture other deficiencies than model error.

\subsubsection{Learning from less frequent snapshots}

Because the true model is chaotic, forecasts are highly sensitive to the initial condition. In addition, the accuracy of forecasts at longer lead times is generally more sensitive to incorrect model parameters, while the accuracy of the training dataset is unchanged. This is beneficial when training the surrogate model with the analysis since it reduces the impact of the analysis error in the training dataset. On the other hand, longer forecast are inherently harder to predict. Therefore, we expect to see a trade-off between both effects when increasing the length of the forecasts.

By construction, the surrogate models are trained to emulate the dynamics over one DAW, \cref{eq:illustration-offline-training-map}. Changing the DAW length $L$ is not optimal, because $L=\num{6}$ minimises the sRMSE and hence it ensures the most accurate analysis on average. Another possibility is to keep $L=\num{6}$ and train the surrogate models to emulate the dynamics over multiple DAWs, \textit{i.e.} to reproduce the map
\begin{equation}
    \label{eq:illustration-offline-training-map-multiple}
    \mathbf{x}^{\mathsf{a}}_{kL}\mapsto\mathbf{x}^{\mathsf{a}}_{\left(k+N\right)L},
\end{equation}
where $N$ is the number of DAWs over which the models should emulate the dynamics. This technique has been used by \citet{farchi-2021} in the context of the RC method, with only partial success. Indeed, using $N>\num{1}$ systematically worsen the DA scores, which is not a surprise because the assumption of linear growth of errors in time is less and less valid for longer forecasts. In this section, we evaluate this technique with the TC method, which does not suffer from the same limitations as the RC method (in particular it does not require additional assumption for DA experiments).

\begin{figure}[tbh]
    \centering
    \includegraphics{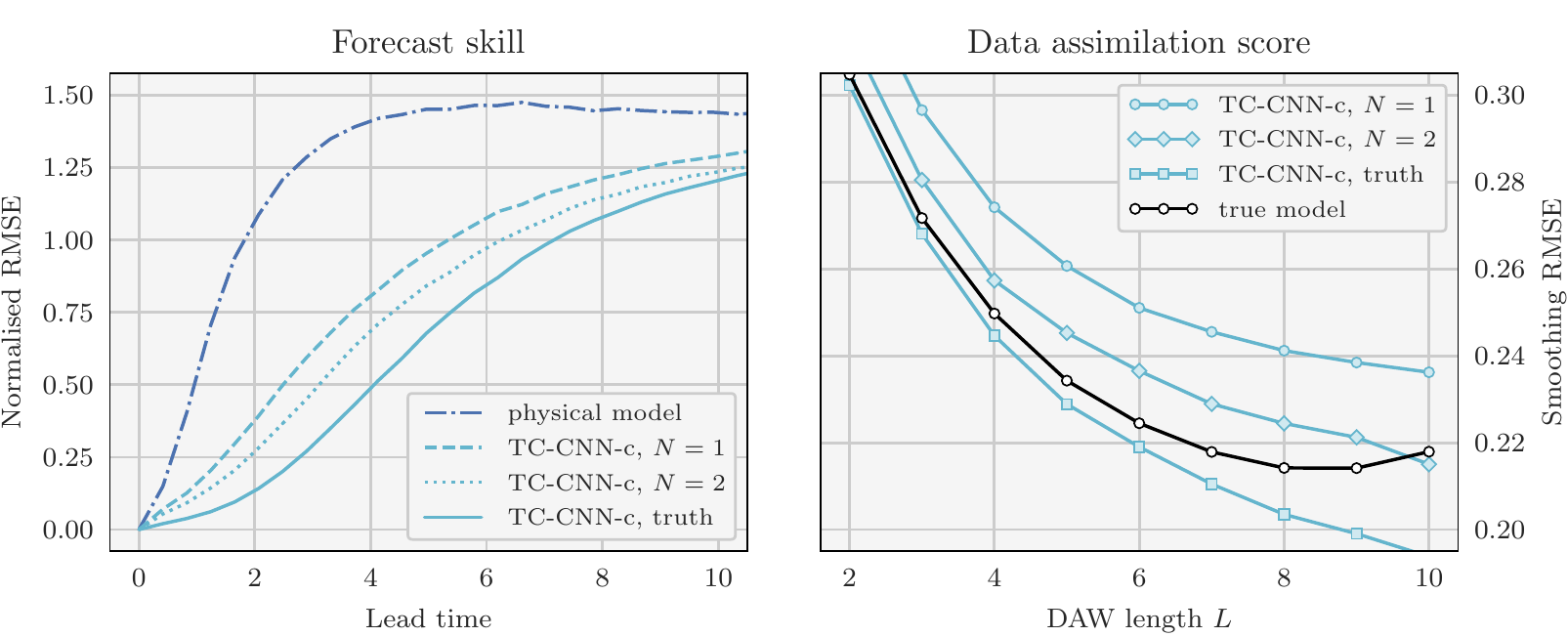}
    \caption{Same as \cref{fig:illustration-offline-models-fs-da} for the physical model (in blue), the true model (in black), and the trained surrogate model TC-CNN-c (in cyan). The surrogate model is trained either with the analysis over one DAW ($N=\num{1}$), over two DAWs ($N=\num{2}$), or with the truth over one DAW ($N=\num{1}$).}
    \label{fig:illustration-offline-fs-da-double}
\end{figure}

\Cref{fig:illustration-offline-fs-da-double} shows the forecast skill and the DA score for TC-CNN-c trained with the analysis using $N=\num{2}$. For comparison, the scores for TC-CNN-c trained with the analysis and with the truth, both using $N=\num{1}$, are taken from \cref{fig:illustration-offline-fs,fig:illustration-offline-da} and reproduced here. These results confirm that increasing the length of the forecasts to predict yields more accurate surrogate models, both for forecast and DA experiments. Moreover, we have checked that the scores get worse when further increasing the length of the forecasts to $N=\num{4}$ (not shown here). This indicates that the trade-off aforementioned reaches an optimum for $N=\num{2}$ or $N=\num{3}$.

\subsubsection{Iterating the DA-ML method}
\label{sssec:illustration-offline-iterating-da-ml}

Throughout the previous experiments, we have seen different performances depending on whether the surrogate models are trained with the analysis or with the truth, the latter case being equivalent to using dense and noiseless observations. Iterating the DA-ML method, as originally proposed by \citet{brajard-2020} and formalised by \citet{bocquet-2020a} as a coordinate descent optimisation, is a way to bring the performances of the surrogate models trained with the analysis closer to that of the surrogate models trained with the truth.

Given the results of the DA experiment with the RC method in \cref{sssec:illustration-offline-corrected-da-exp}, it is possible that an additional iteration of the DA-ML method will not improve the DA score by much. This is not the case with the TC method because in this case, the analysis with the surrogate model is much more accurate than that of the physical (non-corrected) model. However, for the TC method we would like to show an alternative approach in the next section. 

\section{Online learning of model error with tendency correction}
\label{sec:methodology-online}

\subsection{From offline to online learning}

As already mentioned, the DA-ML method presented in \cref{sec:methodology-offline} and illustrated in \cref{sec:illustration-offline} is an offline learning method, meaning that the training starts only once all observations have been assimilated. This makes the method simple and flexible because the ML and DA steps are independent from each other. As a consequence, it is probably easier to extract all the information from the analysis during the ML step.

On the other hand, using an offline approach also has drawbacks. As suggested in \cref{sssec:illustration-offline-iterating-da-ml}, the DA-ML method needs to be iterated to give the best performance. At each iteration, the entire set of observations has to be re-assimilated, which can be problematic when the DA step is numerically expensive (for example with a realistic model). This is particularly concerning in the case of an operational model for which a new correction must be trained each time the model gets updated. Such issues does not affect online approaches, since the training could start as soon as the first observation arrives.

In our context, online learning means that, each time a new batch of observation becomes available, we have to estimate both the state of the system and the surrogate model at the same time. To address this problem, the most natural approach is to use the formalism of DA with an augmented state containing the current state of the system $\mathbf{x}$ and the parameters of the surrogate model $\mathbf{p}$, following the principle formulated by \citet{jazwinski-1970}. The resulting inference problem shares many aspects with classical parameter estimation in DA \citep{ruiz-2013}.

\subsection{A new formulation of weak-constraint 4D-Var}

Following the approach described in \cref{ssec:illustration-offline-da-physical-model}, the observations are assimilated using the 4D-Var algorithm, with consecutive windows of $L$ batches of observations. Since the algorithm now has to estimate the model parameters in addition to the model state, each 4D-Var problem consists in minimising the cost function
\begin{equation}
    \label{eq:methodology-online-cost-function}
    \mathcal{J}\left(\mathbf{p}_{k}, \mathbf{x}_{k}\right) = \frac{1}{2}\left\|\mathbf{p}_{k}-\mathbf{p}^{\mathsf{b}}_{k}\right\|^{2}_{\mathbf{B}^{-1}_{\mathsf{p}}} + \frac{1}{2}\left\|\mathbf{x}_{k}-\mathbf{x}^{\mathsf{b}}_{k}\right\|^{2}_{\mathbf{B}^{-1}_{\mathsf{x}}} + \frac{1}{2}\sum_{l=0}^{L-1}\left\|\mathbf{y}_{k+l}-\mathcal{H}\circ\mathcal{M}_{l\Delta t}\left(\mathbf{p}_{k}, \mathbf{x}_{k}\right)\right\|^{2}_{\mathbf{R}^{-1}},
\end{equation}
where $\mathbf{p}^{\mathsf{b}}_{k}$ and $\mathbf{B}_{\mathsf{p}}$ are the background and background error covariance matrix for model parameters, respectively. Comparing \cref{eq:methodology-online-cost-function} to the 4D-Var cost function without model parameters, \cref{eq:llustration-offline-da-setup-cost-function}, two differences can be highlighted. First, the resolvent of the physical model $\mathcal{M}^{\mathsf{p}}_{l\Delta t}$ is replaced with the resolvent of the surrogate model $\mathcal{M}_{l\Delta t}$, which now depends on the model parameters. Second, a background (or prior) term on model parameters has been added. This background term is very important because, in this cycled context, it carries out the information on model parameters from one DAW to the next. Indeed, the background on model parameters for the next DAW is equal to the analysis on model parameters of the current DAW:
\begin{equation}
    \mathbf{p}^{\mathsf{b}}_{k+L} = \mathbf{p}^{\mathsf{a}}_{k}.
\end{equation}
In other words, the forecast model for model parameters is the persistence, which is consistent with the fact that the model is autonomous. 

Like WC 4D-Var, \cref{eq:methodology-online-cost-function} can be inferred from Bayes' rule, but \cref{eq:methodology-online-cost-function} additionally neglects the cross-covariances between the background errors for model state and for model parameters. Including cross-covariances between model state and model parameters is possible but requires either a prior knowledge or the use of an ensemble to estimate them, which is beyond the scope of the present work.

Our assimilation method is summarised in \cref{alg:methodology-online-wc-4dvar} in a cycled context. Rigorously speaking, this could be seen as a SC method because it includes only one state vector in the control variables. However with our method, by contrast with SC 4D-Var, the model can be updated during the analysis (when the model parameters change). Therefore, considering a broader definition of WC methods, \cref{alg:methodology-online-wc-4dvar} can be seen as a specific formulation of WC 4D-Var.

\begin{algorithm}
\caption{NN formulation of WC 4D-Var in a cycled context}
\label{alg:methodology-online-wc-4dvar}
\begin{algorithmic}[1]
\renewcommand{\algorithmicrequire}{\textbf{Parameters:}}
\renewcommand{\algorithmicensure}{\textbf{Input:}}
\renewcommand{\algorithmiccomment}[1]{\hfill$\triangleright$~\textit{#1}}
\REQUIRE{Observation operator $\mathcal{H}$, surrogate model $\mathcal{M}$ with NN, background error covariance matrices for model state and model parameters $\mathbf{B}_{\mathsf{x}}$ and $\mathbf{B}_{\mathsf{p}}$, observation error covariance matrix $\mathbf{R}$, DAW length $L$}
\ENSURE{Initial background for model state and model parameters $\mathbf{x}^{\mathsf{b}}_{0}$ and $\mathbf{p}^{\mathsf{b}}_{0}$, observations $\left\{\mathbf{y}_{0},\ldots,\mathbf{y}_{\left(N_{\mathsf{t}}+1\right)L}\right\}$}
\FOR{$k=0$ \TO $N_{\mathsf{t}}$}
\STATE{Compute the analysis $\mathbf{p}^{\mathsf{a}}_{kL}$ and $\mathbf{x}^{\mathsf{a}}_{kL}$ by minimising $\mathcal{J}\left(\mathbf{p}_{kL}, \mathbf{x}_{kL}\right)$, \cref{eq:methodology-online-cost-function}}\COMMENT{analysis at time $t_{kL}$}
\STATE{Forecast the model parameters $\mathbf{p}^{\mathsf{b}}_{\left(k+1\right)L}=\mathbf{p}^{\mathsf{a}}_{kL}$}\COMMENT{using persistence}
\STATE{Forecast the model state $\mathbf{x}^{\mathsf{b}}_{\left(k+1\right)L}=\mathcal{M}_{L\Delta t}\left(\mathbf{p}^{\mathsf{a}}_{kL}, \mathbf{x}^{\mathsf{a}}_{kL}\right)$}\COMMENT{using the surrogate model}
\ENDFOR
\RETURN{$\left(\mathbf{x}^{\mathsf{a}}_{kL}, \mathbf{p}^{\mathsf{a}}_{kL}\right)$ for $k\in\left\{0,\ldots,N_{\mathsf{t}}\right\}$}\COMMENT{analysis trajectory}
\end{algorithmic}
\end{algorithm}

Optionally, the model parameters can be pre-trained. For completeness, we would like to mention the similarities with the algorithms developed by \citet{bocquet-2020a,malartic-2021} which solve the same kind of problems but in a filtering context (\textit{i.e.} with $L=\num{1}$) with several variants of the ensemble Kalman filter. Finally note that, just as the DA-ML method of \cref{sec:methodology-offline} can be used for model error correction instead of full model emulation, the present formulation of WC 4D-Var can be used for model error correction as well.

\section{Numerical illustration with the two-scale Lorenz model (part II)}
\label{sec:illustration-online}

In this section, we illustrate the online learning method developed in \cref{sec:methodology-online} using the same model error setup as in \cref{sec:illustration-offline}. The true model is the L05III model and the physical model is the L96 model. However, because the online learning approach is based on the sole formalism of DA, we only use the surrogate models TC-CNN-b and TC-CNN-c, built with the TC method, since we have shown in \cref{sssec:illustration-offline-corrected-da-exp} that it is the most consistent and the most efficient for DA experiments.

\subsection{Data assimilation setup}

For this illustration, we take the same DA problem as in \cref{ssec:illustration-offline-da-physical-model}. The truth is generated using the true model, and observations are taken every $\Delta t=\num{0.05}$ from the slow variables only, using \cref{eq:illustration-offline-da-observations}.

We start the experiments by assimilating the observations using the SC 4D-Var algorithm with the exact same setup as in \cref{ssec:illustration-offline-da-physical-model}. In particular, we use the physical model, and we choose $L=\num{6}$ as it yields the best results with this model. After a total of $\num{1024}$ DA cycles, which is enough to ensure the convergence of the analysis error statistics, we shift to the NN formulation of WC 4D-Var (\cref{alg:methodology-online-wc-4dvar}). In other words, we switch on model error correction. For the following cycles, we keep $L=\num{6}$ and $\mathbf{B}_{\mathsf{x}}=b_{\mathsf{x}}^{2}\mathbf{I}$, although $b_{\mathsf{x}}$ can be different than in the $\num{1024}$ preliminary cycles. In addition, we set $\mathbf{B}_{\mathsf{p}}=b_{\mathsf{p}}^{2}\mathbf{I}$, where $b_{\mathsf{p}}$ is another algorithmic parameter to specify. Finally for consistency, the initial background for model parameters $\mathbf{p}^{\mathsf{b}}_{0}$ is constructed using the NN initialisation method described in \cref{sssec:illustration-offline-nn-initialisation}.

With WC 4D-Var, at each cycle the cost function $\mathcal{J}$, \cref{eq:methodology-online-cost-function}, is minimised using the same L-BFGS algorithm as for the SC 4D-Var. The gradient of $\mathcal{J}$, with respect to both model state and model parameters, is computed exactly using automatic differentiation, and the starting point of the minimisation is $\left(\mathbf{p}^{\mathsf{b}}_{k}, \mathbf{x}^{\mathsf{b}}_{k}\right)$. The accuracy of the analysis is measured using the RMSE \emph{on model state} (analysis minus truth) at the start of the DAW, which corresponds to the sRMSE defined in \cref{ssec:illustration-offline-da-physical-model}. In this cycled context, the sRMSE only improves if the model is getting more accurate. Nevertheless, we also measure the accuracy of the model by computing the tMSE defined in \cref{sssec:illustration-offline-training-surrogate-models}.

Finally note that the $\num{1024}$ preliminary cycles with the physical model are not mandatory. We have checked that the results are qualitatively similar without them. In fact, adding these preliminary cycle is a way to get in real condition, where we need to correct a physical model which is already running since a while. In the following section, we do not discuss the preliminary cycles.

\subsection{Results with TC-CNN-b as surrogate model}
\label{ssec:illustration-online-tc-cnn-b}

Let us start by applying the method to TC-CNN-b. In other words, in this case $\mathcal{M}_{l\Delta t}$ in \cref{eq:methodology-online-cost-function} is the resolvent of TC-CNN-b for an integration of $l\Delta t$. For this experiment, we use the following algorithmic parameters, which we have empirically found to yield good performances:
\begin{subequations}
    \label{eq:illustration-online-tc-cnn-b-parameters}
    \begin{align}
        b_{\mathsf{x}} &= 0.28 + 0.15 \times \exp\left(-t/256\right), \\
        \widehat{b_{\mathsf{p}}} &= 0.001 + 0.1\times\exp\left(-t/1024\right), \\
        b_{\mathsf{p}} &= \min\left[\,0.05, \widehat{b_{\mathsf{p}}}\,\right],
    \end{align}
\end{subequations}
where $t$ is the time measured in number of DAWs after the $\num{1024}$ preliminary cycles. The rationale behind this choice is that the optimal values of $b_{\mathsf{x}}$ and $b_{\mathsf{p}}$ should be larger at the start of the experiment than at the end (when the model is more accurate). To come up with \cref{eq:illustration-online-tc-cnn-b-parameters}, we first chose the shape of $b_{\mathsf{x}}$ and $b_{\mathsf{p}}$ (exponential decay) and we then tuned the coefficients until we got satisfying results. Also note that we have checked that the results are qualitatively similar when replacing the exponential decay of $b_{\mathsf{x}}$ and $b_{\mathsf{p}}$ with a linear decay.

\begin{figure}[tbh]
    \centering
    \includegraphics{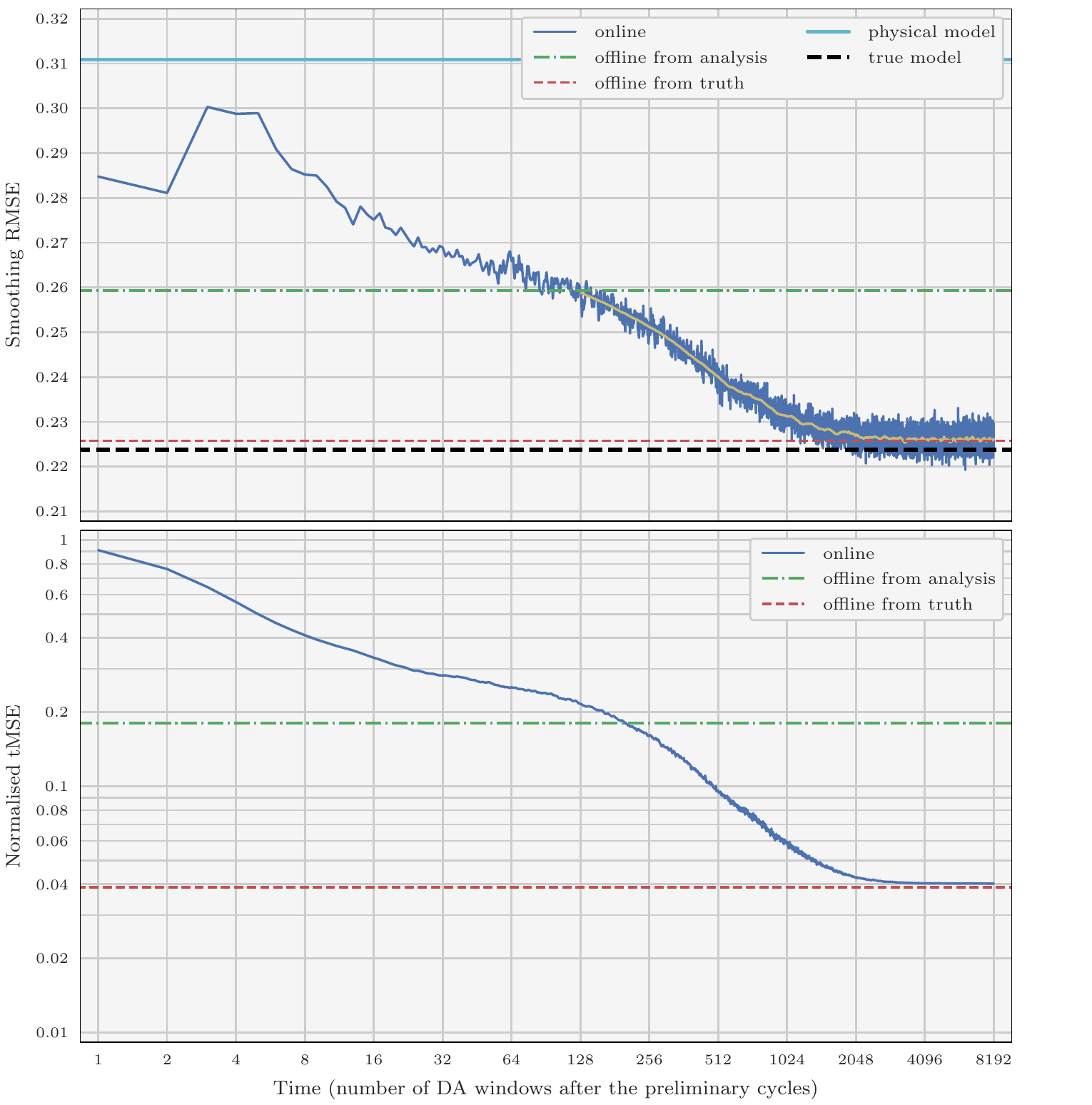}
    \caption{Time series of sRMSE (top panel) and tMSE (bottom panel) for the online experiment with TC-CNN-b (in blue). The tMSE is computed using a test dataset of size $N_{\mathsf{t}}=1024$ and then normlised by the tMSE of the physical model. Both sRMSE and tMSE are averaged over $\num{512}$ repetitions of the experiment. In addition, the yellow line shows the moving-average of the sRMSE over $\num{128}$ DAWs. For comparison, the horizontal lines show the scores for the physical model (in cyan), the true model (in black), TC-CNN-b trained offline with the analysis (in green) and trained offline with the truth (in red). }
    \label{fig:illustration-online-cnnb}
\end{figure}

\Cref{fig:illustration-online-cnnb} shows the time series of sRMSE and tMSE throughout the experiment, after the $\num{1024}$ preliminary cycles. For comparison, the scores obtained when TC-CNN-b is trained using the offline DA-ML approach, both with the analysis and with the truth, are taken from \cref{fig:illustration-offline-models-tmse,fig:illustration-offline-da} and reproduced here. First of all, the experiment is a success: the algorithm is working as expected and steadily improves the model, which can be seen both in the sRMSE (DA score) and in the tMSE (forecast score). After $\num{128}$ to $\num{256}$ cycles the model becomes more accurate than if trained offline with the analysis. Finally, the accuracy converges after $\num{2048}$ to $\num{4096}$ cycles. In the end, the model is almost as accurate as if trained offline with the truth! This is a strong result because, as mentioned in \cref{sssec:illustration-offline-training-surrogate-models}, training with the truth illustrates the full potential of a surrogate model. This shows that our online learning method has been able to extract all the information from the observations, and that we cannot expect better results with this surrogate model.

\subsection{Results with TC-CNN-c as surrogate model}
\label{ssec:illustration-online-tc-cnn-c}

We now apply the method to TC-CNN-c. For this experiment, the algorithmic parameters, once again chosen on empirical grounds, are slightly different:
\begin{subequations}
    \label{eq:illustration-online-tc-cnn-c-parameters}
    \begin{align}
        b_{\mathsf{x}} &= 0.26 + 0.20 \times \exp\left(-t/256\right), \\
        \widehat{b_{\mathsf{p}}} &= 0.01 + 0.05\times\exp\left(-t/3072\right), \\
        b_{\mathsf{p}} &= \min\left[\,0.05, \widehat{b_{\mathsf{p}}}\,\right],
    \end{align}
\end{subequations}
with $t$ being once again the time measured in number of DAWs after the $\num{1024}$ preliminary cycles.

\begin{figure}[tbh]
    \centering
    \includegraphics{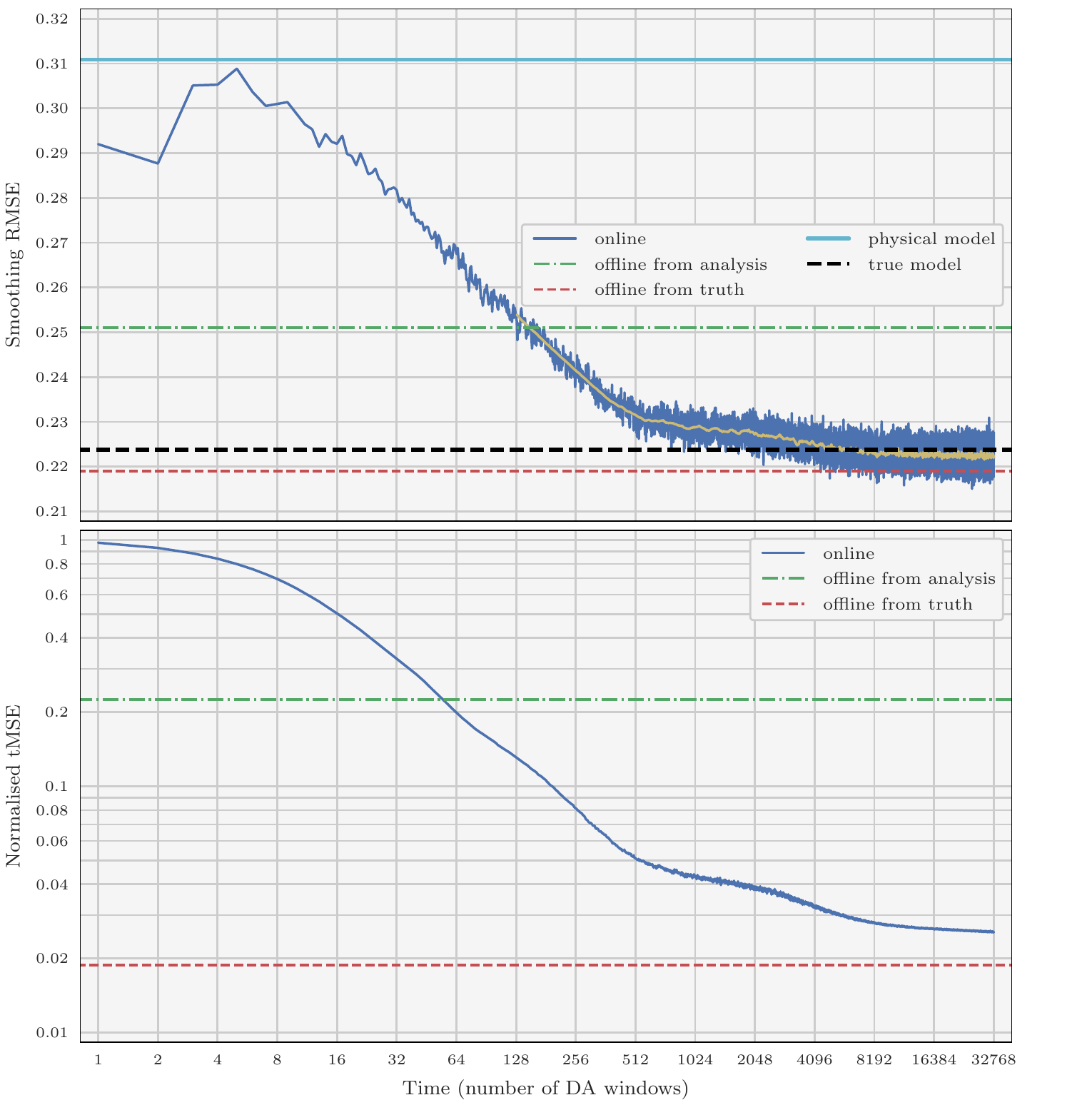}
    \caption{Same as \cref{fig:illustration-online-cnnb} for the online experiment with TC-CNN-c.}
    \label{fig:illustration-online-cnnc}
\end{figure}

\Cref{fig:illustration-online-cnnc} shows the time series of sRMSE and tMSE throughout the experiment, after the $\num{1024}$ preliminary cycles. The success of the experiment is as clear as with TC-CNN-b: the algorithm steadily improves the model, which can be seen both in the sRMSE and in the tMSE. After $\num{128}$ to $\num{256}$ cycles the model becomes more accurate than if trained offline with the analysis. However, even though the total number of DA cycles increases, the accuracy has not yet converged at the end of the experiment. One must keep in mind that the correction provided by CNN-c is here nonlinear, contrary to the previous experiment with TC-CNN-b where the correction provided by CNN-b is linear. Therefore, we think that the present experiment is a good illustration of the increased complexity of training nonlinear NNs. Nevertheless, the accuracy of the model after $\num{8192}$ is remarkable and in particular the sRMSE is lower than when using the true model! We also think that, if we were to extend the experiment with an appropriate tuning for $b_{\mathsf{p}}$, the model would in the end be almost as accurate as if trained offline with the truth, just as in the previous experiment with TC-CNN-b.

\subsection{Additional remarks on the online experiments}

While preparing the online experiments, we have found that tuning the algorithmic parameter $b_{\mathsf{p}}$ is very important. If $b_{\mathsf{p}}$ is too small, the algorithm gives too much weight to the background on model parameters $\mathbf{p}^{\mathsf{b}}_{k}$. Even if this does not stop the learning process, it tapers the model parameter update, which makes the convergence slower\footnote{The convergence speed is measured here in number of DA cycles before convergence and not in wall-clock execution time.}. On the other hand if $b_{\mathsf{p}}$ is too large, the algorithm overfits the model parameters to the observation window, which can yield divergence. As mentioned in \cref{ssec:illustration-online-tc-cnn-b}, what makes the tuning of $b_{\mathsf{p}}$ really complex here is that, as the model steadily improves, the optimal value of $b_{\mathsf{p}}$ decrease. The values selected for the experiments, \cref{eq:illustration-online-tc-cnn-b-parameters,eq:illustration-online-tc-cnn-c-parameters}, have been chosen by \emph{trial and error}. Even though they yield good performances, they have not been optimally tuned. This means that a faster convergence could have been most likely obtained with other values. Finally, note that the algorithmic parameter $b_{\mathsf{p}}$ here is very similar to the tapering parameter introduced by \citet{bocquet-2020a, malartic-2021} in their variants of the ensemble Kalman filter.

The online experiments use the zero/random NN initialisation method described in \cref{sssec:illustration-offline-nn-initialisation}. Therefore, at the start of the WC 4D-Var cycles the surrogate model is equivalent to the physical model. Other initialisation methods are possible. For example, one can use the set of parameters obtained after the offline learning method of \cref{ssec:illustration-offline-rc-tc}. We have implemented this method (note illustrated here) and checked that, with an appropriate tuning of $b_{\mathsf{p}}$, the final scores are the same than with the zero/random initialisation but the convergence is somewhat faster.

Finally, the online experiments use the same DAW length as with the physical model, $L=\num{6}$. However, since the accuracy of the model increases during the experiment, increasing $L$ is a reasonable option. We have performed the online experiments with $L=\num{10}$ (not illustrated here). The evolution of the tMSE (forecast score) is very similar to that with $L=\num{6}$, but, as expected from \cref{fig:illustration-offline-da}, the sRMSE (DA score) is close to $\num{0.2}$, significantly lower than with $L=\num{6}$.

\subsection{Tendency correction in a realistic model}

Both the offline and online learning experiments illustrate the efficiency of the TC method. Including a TC into a complex realistic model is not immediate, depending on the structure of the code, because the correction term must be added to the code computing the tendencies. In order to use the variational methods illustrated in the present work to train a TC, both offline and online, we must be able to compute the gradient of the resolvent of the total model (physical model with correction) with respect to the model state and to the model parameters. As we have shown in \cref{ssec:methodology-offline-univariate-example}, these gradients depend on the gradients of the correction term, which can be easily obtained with a ML library, but also on the TL operator of the physical model. The present work, with a simple model, has been largely facilitated by the fact that the code of the physical model has been written entirely with a ML library, which implies that we have benefited from automatic differentiation. For realistic models, which are inherently more complex, it is essential to have efficient differentiation methods.

Beyond these technical aspects, the number of trainable model parameters is a potential source of concern. In the present work, we have tried to keep it as small as possible, and ended up with $\num{113}$ parameters (for both CNN-b and CNN-c). In preliminary experiments, a much larger NN (a fully-connected NN with a total of $\num{36x64}+\num{64}+\num{64x36}+\num{36}=\num{4708}$ parameters) has been tested and we found similar performances, although with more training data. Even though $\num{113}$ parameters (or even $\num{4708}$) will most likely not be enough to correct a complex, realistic model, we have the hope that with smart ML models, we will be able to keep the number of trainable parameters under control \citep{bonavita-2020}.

\subsection{Generalisation to non-autonomous dynamics}
\label{ssec:illustration-online-generalisation-non-autonomous}

We conclude this test series by briefly discussing the possibility to extend the present work to non-autonomous dynamics. With an offline learning method, a non-autonomous dynamics can only be learnt if (i) the time-dependency of the model is parametrised (which implies that time should be among the set of predictors) and (ii) the training dataset is large enough to infer the parametrisation. With an online learning method, parametrising the time-dependency is also an option, but such parametrisation is not mandatory if the time evolution of the dynamics is slow. For example, the online experiments of \cref{ssec:illustration-online-tc-cnn-b,ssec:illustration-online-tc-cnn-c} would probably also work if the dynamics evolution is no faster than a few hundred DA cycles.

\section{Conclusions}
\label{sec:conclusions}

Combining DA and ML to emulate a dynamical model has been originally proposed by \citet{brajard-2020}. The use of DA is essential here to assimilate sparse and noisy observations, which cannot be rigorously treated with ML methods alone. The same strategy can be used for model error correction instead of full model emulation. This has many advantages as it makes the inference problem easier \citep{jia-2019,watson-2019}. In practice, a correction term can be included either in the model resolvent (\textit{i.e.} as an integrated term between two forecast times) or directly in the model tendencies \citep{bocquet-2019a, farchi-2021}. 

In the present article, we have compared the two methods. The first method, which we have called RC (resolvent correction), is easy to implement and has already been illustrated using low-order models by \citep{brajard-2020b, farchi-2021}. The second method, which we have called TC (tendency correction), is more technical to implement, in particular because it requires the adjoint of the physical model to correct, but it has the advantage of being more flexible, in particular for short-term forecasts. Both methods have been tested using the two-scale Lorenz system. In this case, model error primarily comes from unresolved small-scales processes (the fast variables). We have used noisy observations of the slow variables to build three different surrogate models. All three surrogate models are hybrid, with a physical part, the non-corrected model, and a statistical part, the correction term, built using simple NNs. The first surrogate model uses the RC method, while the other two use the TC method. The surrogate models are trained using the offline DA-ML method of \citet{brajard-2020} and then evaluated in forecast and DA experiments. The results show that with the TC method, the surrogate models benefit from the interaction between the physical model and the NN, in such a way that it is possible to use much smaller NNs and much fewer training data to get similar results. The accuracy in forecast experiments is somewhat similar between the models using the RC and the TC method. By contrast, the models using the TC method significantly outperform the models using the RC method in DA experiments. This can be explained by the violation of the assumption of linear error growth in time, necessary with the RC method for the short-term forecasts within each DA experiment.

In the second part of this paper, we explored the possibility to train the surrogate models in an online fashion. With sparse and noisy observations, this means that we would have to learn both model state and model parameters at the same time. To address this problem, we introduce a new DA algorithm, which can be seen as a new formulation of WC 4D-Var. The algorithm has been implemented and tested using the same model error setup with the two-scale Lorenz system. The results show that online learning works as expected: the surrogate model is steadily improved, which can be seen both in the DA score and the forecast score; it quickly becomes more accurate than if trained offline with the analysis. At the end of the experiment, the model is almost as accurate as if trained offline with the truth. These results show that with online learning, it is possible to extract all the information from sparse and noisy observations.  

Even though the results of the online experiments are promising, the NN formulation of WC 4D-Var derived in the present work could still benefit from methodological developments. In our experiments, we have found that tuning the algorithmic parameter $b_{\mathsf{p}}$ is a difficult but critical task. Ideally, the tuning method should be adaptive, with the objective of making the convergence faster and more accurate \citep{bocquet-2020a}. Furthermore, the method implicitly assumes that background errors for model state and model parameters are uncorrelated. Including cross-correlations between model state and model parameters is possible; it would be interesting to check whether this could make a difference in the accuracy of the analysis. More generally, the conclusions of the present work, and in particular the advantages of the TC method over the RC method, must be confirmed using higher-dimensional models.

\section*{Acknowledgements}

The authors are grateful to an anonymous reviewer for insightful comments which helped improve the manuscript. CEREA is a member of Institut Pierre--Simon Laplace.

\bibliography{bibtex}

\end{document}